\documentclass[11pt,a4paper]{article}
\usepackage{authblk}
\usepackage[hyperref]{emnlp2018}
\usepackage{times}
\usepackage{latexsym}
\usepackage{graphicx}
\usepackage[textsize=small,disable]{todonotes}
\usepackage[textsize=small,disable]{todonotes}
\usepackage{amsmath}
\usepackage{xcolor}
\usepackage{booktabs}
\usepackage{subcaption}

\usepackage{tikz}
\usetikzlibrary{calc,trees,positioning,arrows,chains,shapes.geometric,%
  decorations.pathreplacing,decorations.pathmorphing,shapes,%
  matrix,shapes.symbols,fit,decorations,arrows.meta}

\newcommand{\eg}{\emph{e.g.}}
\newcommand{\ie}{\emph{i.e.}}

\newcommand{\quoteYes}{``Yes''}
\newcommand{\quoteNo}{``No''}

\newcommand{\dsacr}{ShARC}
\newcommand{\dsfull}{Shaping Answers with Rules through Conversation}

\usepackage{cleveref} 
\usepackage{microtype} 
\usepackage{amssymb}
\usepackage{amsmath}
\usepackage{bm}
\usepackage{wrapfig}
\usepackage{enumitem}
\usepackage{xcolor}
\usepackage{float}

\definecolor{nice-red}{HTML}{E41A1C}
\colorlet{dark-red}{nice-red!80!black}
\definecolor{nice-orange}{HTML}{FF7F00}
\colorlet{dark-orange}{orange!85!black}
\definecolor{nice-yellow}{HTML}{FFC020}
\definecolor{nice-green}{HTML}{4DAF4A}
\definecolor{nice-blue}{HTML}{377EB8}
\definecolor{nice-purple}{HTML}{984EA3}

\def\itodo{\textbf{\color{red}TODO}}

\newcommand{\Seb}[1]{\todo[inline,backgroundcolor=orange!20!white]{Sebastian: #1}}
\newcommand{\SebMargin}[1]{\todo[backgroundcolor=orange!20!white]{Sebastian: #1}}

\newcommand{\Max}[1]{\todo[inline,backgroundcolor=blue!20!white]{Max: #1}}
\newcommand{\MaxMargin}[1]{\todo[backgroundcolor=blue!20!white]{Max: #1}}

\newcommand{\Patrick}[1]{\todo[inline,backgroundcolor=blue!20!white]{Pat: #1}}
\newcommand{\PatrickMargin}[1]{\todo[backgroundcolor=blue!20!white]{Pat: #1}}

\newcommand{\MikeMargin}[1]{\todo[backgroundcolor=yellow!20!white]{Mike: #1}}

\newcommand{\tim}[1]{\todo[backgroundcolor=red!20!white]{Tim: #1}}
\newcommand{\sameer}[1]{\todo[backgroundcolor=purple!40!white]{sameer: #1}}

\usepackage{url}

\aclfinalcopy % Uncomment this line for the final submission

\newcommand{\subsubsect}[1]{\paragraph{#1}}

\newcommand*{\affaddr}[1]{#1} % No op here. Customize it for different styles.
\newcommand*{\affmark}[1][*]{\textsuperscript{#1}}
\newcommand*{\email}[1]{\texttt{#1}}

\title{Interpretation of Natural Language Rules in \\Conversational Machine Reading}

\author{%
Marzieh Saeidi\affmark[1]\thanks{\ \ These three authors contributed equally} , Max Bartolo\affmark[1]\normalfont\textsuperscript{*}\textbf{,} \textbf{Patrick Lewis\affmark[1]}\normalfont\textsuperscript{*}\textbf{,} \textbf{Sameer Singh\affmark[1,2]}\textbf{,}\authorcr Tim Rockt{\"a}schel\affmark[3], Mike Sheldon\affmark[1], Guillaume Bouchard\affmark[1], 
and Sebastian Riedel\affmark[1,3]\\

\affaddr{\affmark[1]Bloomsbury AI}\\
\affaddr{\affmark[2]University of California, Irvine}\\
\affaddr{\affmark[3]University College London}\\
\\
\email{\{marzieh.saeidi,maxbartolo,patrick.s.h.lewis\}@gmail.com}\\
}

\date{}

\begin{document}
\maketitle
\begin{abstract}
Most work in machine reading focuses on question answering problems where the answer is directly expressed in the text to read. However, many real-world question answering problems require the reading of text not because it contains the literal answer, but because it contains a recipe to derive an answer together with the reader's background knowledge. One example is the task of interpreting regulations to answer ``Can I...?'' or ``Do I have to...?'' questions such as ``I am working in Canada. Do I have to carry on paying UK National Insurance?'' after reading a UK government website about this topic. This task requires both the interpretation of rules and the application of background knowledge. It is further complicated due to the fact that, in practice, most questions are underspecified, and a human assistant will regularly have to ask clarification questions such as ``How long have you been working abroad?'' when the answer cannot be directly derived from the question and text. 
In this paper, we formalise this task and develop a crowd-sourcing strategy to collect 32k task instances based on real-world rules and crowd-generated questions and scenarios. We analyse the challenges of this task and assess its difficulty by evaluating the performance of rule-based and machine-learning baselines. We observe promising results when no background knowledge is necessary, and substantial room for improvement whenever background knowledge is needed. 
\end{abstract}

\section{Introduction}
\begin{figure}[h!]
\centering
\includegraphics[width=0.50\textwidth,angle=0,clip,trim={192 42 295 310}]{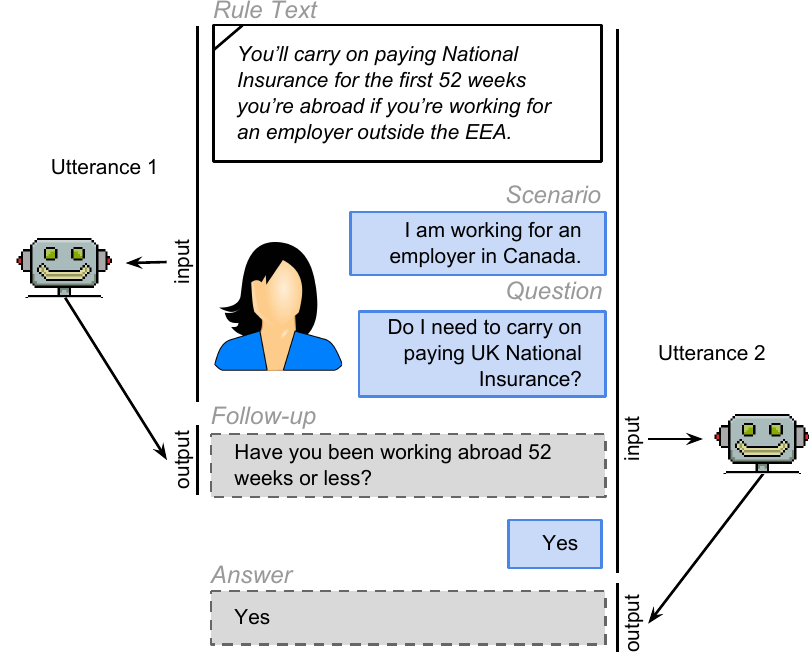}
\caption{\label{example} An example of two utterances for rule interpretation. In the first utterance, a follow-up question is generated. In the second, the scenario, history and background knowledge (Canada is not in the EEA) is used to arrive at the answer \quoteYes{}.}
\end{figure}

% Motivate the interpretation of rules
% Are there some interesting citations for this paragraph?
% Maybe: Natural Language Processing of Rules and Regulations for Compliance in the Cloud

% QA requires rules
There has been significant progress in teaching machines to read text and answer questions when the answer is directly expressed in the text~\cite{Rajpurkar2016_SQUAD,Joshi_2017_TriviaQA,DBLP:journals/corr/abs-1710-06481,hermann2015teaching}. \SebMargin{Can we add some old school extractive QA as well? Any links?}  However, in many settings, the text contains rules expressed in natural language that can be used to infer the answer when combined with background knowledge, rather than the literal answer. For example, to answer someone's question ``I am working for an employer in Canada. Do I need to carry on paying National Insurance?'' with \quoteYes{}\sameer{check: is Germany/No better example?}, one needs to read that ``You'll carry on paying National Insurance if you're working for an employer outside the EEA'' and understand how the rule and question determine the answer.    
   
% QA requires clarification questions
Answering questions that require rule interpretation is often further complicated due to missing information in the question. For example, as illustrated in \Cref{example} (Utterance 1), the actual rule also mentions that National Insurance only needs to be paid for the first 52 weeks when abroad. This means that we cannot answer the original question without knowing how long the user has already been working abroad. Hence, the correct response in this conversational context is to issue another query such as ``Have you been working abroad 52 weeks or less?''      
   
% Introduce the task  
To capture the fact that question answering in the above scenario requires a dialog, we hence consider the following \emph{conversational machine reading}~(CMR) problem as displayed in \Cref{example}: Given an input question, a context scenario of the question, a snippet of supporting rule text containing a rule, and a history of previous follow-up questions and answers, predict the answer to the question \MikeMargin{All `yes' and `no's after the introduction are formatted differently, should these be the same?} (\quoteYes or \quoteNo) or, if needed, generate a follow-up question whose answer is necessary to answer the original question.
Our goal in this paper is to create a corpus for this task, understand its challenges, and develop initial models that can address it. 
% describing the question context

% The high level corpus protocol
\sameer{very much so}\SebMargin{Is this better now?}
To collect a dataset for this task, we could give a textual rule to an annotator and ask them to provide an input question, scenario, and dialog in one go. This poses two problems. First, this setup would give us very little control. For example, users would decide which follow-up questions become part of the scenario and which are answered with \quoteYes{} or \quoteNo{}. Ultimately, this can lead to bias because annotators might tend to answer \quoteYes{}, or focus on the first condition. Second, the more complex the task, the more likely crowd annotators are to make mistakes. To mitigate these effects, we aim to break up the utterance annotation as much as possible. 

We hence develop an annotation protocol in which annotators collaborate with virtual users---agents that give system-produced answers to follow-up questions---to incrementally construct a dialog based on a snippet of rule text and a simple underspecified initial question (e.g., ``Do I need to ...?''), and then produce a more elaborate question based on this dialog (e.g., ``I am ... Do I need to...?''). By controlling the answers of the virtual user, we control the ratio of ``Yes'' and ``No'' answers. And by showing only subsets of the dialog to the annotator that produces the scenario, we can control what the scenario is capturing. The question, rule text and dialogs are then used to produce utterances of the kind we see in \Cref{example}. Annotators show substantial agreement \SebMargin{Explain what agreement means in this dialog} when constructing dialogs with a three-way annotator agreement at a Fleiss' Kappa level of $0.71$.\footnote{This is well within the range of what is considered as substantial agreement~\cite{artstein2008inter}.}
\MaxMargin{Not sure what to do in this TODO. Not sure if this line is necessary.} Likewise, we find that our crowd-annotators produce questions that are coherent with the given dialogs with high accuracy. \sameer{(\itodo{}).}
% We also verify that the generated scenarios have an \MaxMargin{I don't understand what scenario accuracy is} accuracy of almost 0.9, based on our human validation.
% \sameer{Put summary of E2E here?}

% How we are solving the task
In theory, the task could be addressed by an end-to-end neural network that encodes the question, history and previous dialog\MaxMargin{aren't previous dialog and history the same thing? Also, what about scenario?}, and then decodes a Yes/No answer or question. In practice, we test this hypothesis using a seq2seq model \cite{sutskever2014sequence, cho2014learning}, with and without copy mechanisms \MaxMargin{I don't believe we give any results for models without copy mech. That's fine, right?} \PatrickMargin{re:above, I hope its fine, as we dont have results prepared for models without copy mechanisms}\cite{gu2016copy} to reflect how follow-up questions often use lexical content from the rule text. 
We find that despite a training set size of 21,890 training utterances, successful models for this task need a stronger inductive bias due to the inherent challenges of the task: interpreting natural language rules, generating questions, and reasoning with background knowledge. We develop heuristics that can work better in terms of identifying what questions to ask, but they still fail to interpret scenarios correctly. To further motivate the task, we also show in oracle experiments that a CMR system can help humans to answer questions faster and more accurately.

% As first step into this direction, we develop a ``recursive QA'' architecture in which questions are answered by generating and directing follow-up questions either to an internal ``helper QA'' machine that answers the follow-up question automatically, or, if that is not possible, to the user. 
% Human and machine answers are recorded and used to derive the final answer.     

% a ``Master QA'' module answers the user question with a follow-up question (such as) that it directs to a ``Helper QA'' module. This module  
% back to the user, or              
% We compare a series of baseline models and find that seq2seq models achieve relatively high accuracy in predicting whether the next dialog act is a ``Yes'', ``No'', ``Irrelevant'' or a follow-up question, but perform poorly in terms of generating a follow-up question when needed. A heuristic that uses bullet point text and text between conjunctions as questions, combined with a neural model to decide when to ask and when to answer, achieves substantial improvements at a BLEU score of 0.14 (order=$4$) and three-way classification accuracy at $60.5\%$. This suggests ample room for improvement, for example, through the use of pointer-network based architectures~\cite{DBLP:journals/corr/GulcehreANZB16}.

This paper makes the following contributions:
\begin{enumerate}[nosep]
\item We introduce the task of conversational machine reading \tim{contrast with FAIR's ParlAI and NIPS paper: Dialog-based language learning} and provide evaluation metrics\SebMargin{We don't do much in terms of metrics. Maybe de-emphasize}.
\item We develop an annotation protocol to collect annotations for conversational machine reading, suitable for use in crowd-sourcing platforms such as Amazon Mechanical Turk.   
\item We provide a corpus of over 32k conversational machine reading utterances, from domains such as grant  descriptions, traffic laws and benefit programs, and include an analysis of the challenges the corpus poses.      
\item We develop and compare several baseline models for the task and subtasks. % we identified. 
\end{enumerate}

% \sameer{unnecessary paragraph}
% In the following we will first introduce the task, discuss our dataset collection protocol and then describe and analyse the collected dataset in more detail. We will then evaluate baselines, discuss limitations of this work, and conclude.   

\section{Task Definition}\label{section_task}

\newcommand{\support}{\ensuremath{r}}
\newcommand{\history}{\ensuremath{h}}
\newcommand{\word}{\ensuremath{w}}
\newcommand{\x}{\ensuremath{x}}
\newcommand{\question}{\ensuremath{q}}
\newcommand{\scenario}{\ensuremath{s}}
\newcommand{\lensupport}{\ensuremath{n_\support}}
\newcommand{\lenquestion}{\ensuremath{n_\question}}
\newcommand{\lenhistory}{\ensuremath{n_\history}}
\newcommand{\yes}{\ensuremath{\textsc{Yes}}}
\newcommand{\no}{\ensuremath{\textsc{No}}}
\newcommand{\more}{\ensuremath{\textsc{More}}}
\newcommand{\irrelevant}{\ensuremath{\textsc{Irrelevant}}}
\newcommand{\followupquestion}{\ensuremath{\textsc{Follow-up Question}}}
\newcommand{\ruletext}{\ensuremath{\textsc{Rule Text}}}
\newcommand{\utterance}{\ensuremath{\textsc{Utterance}}}
\newcommand{\answer}{\ensuremath{y}}
\newcommand{\vocab}{\ensuremath{W}}
\newcommand{\entailment}{\ensuremath{\textsc{Entailment}}}
\newcommand{\contradiction}{\ensuremath{\textsc{Contradiction}}}
\newcommand{\neutral}{\ensuremath{\textsc{Neutral}}}

Figure~\ref{example} shows an example of a conversational machine reading problem. A user has a question that relates to a specific rule or part of a regulation, such as ``Do I need to carry on paying National Insurance?''. In addition, a natural language description of the context or \emph{scenario}, such as ``I am working for an employer in Canada'', is provided. The question will need to be answered using a small snippet of supporting \emph{rule text}. 
Akin to machine reading problems in previous work~\cite{Rajpurkar2016_SQUAD,hermann2015teaching}, we assume that this snippet is pre-identified. 
We generally assume that the question is \emph{underspecified}, in the sense that the question often does not provide enough information to be answered directly. However, an agent can use the supporting rule text to infer what needs to be asked in order to determine the final answer. In \Cref{example}, for example, a reasonable follow-up question is ``Have you been working abroad 52 weeks or less?''.\sameer{something about that is the adequate question (versus longer dialogs?)}

We formalise the above task on a per-utterance basis. A given dialog corresponds to a sequence of prediction problems, \tim{I don't get the second half of this sentence} one for each utterance the system needs to produce. 
Let \vocab{} be a vocabulary. Let $\question=\word_1 \ldots \word_{\lenquestion}$ be an \emph{input question} and $\support=\word_1 \ldots \word_{\lensupport}$  an input \emph{support} rule text, where $w_i\in W$ is a word from a vocabulary. 
Furthermore, let $\history=(f_1,a_1)\ldots (f_{\lenhistory}, a_{\lenhistory})$ be a dialog \emph{history} where each $f_i \in W^*$ is a \emph{follow-up question}, and each $a_i\in \{ \yes,\no \}$ is a \emph{follow-up answer}. 
Let $\scenario$ be a scenario describing the context of the question. We will refer to $\x = (\question, \support, \history, \scenario)$ as the \emph{input}. Given an input \x, our task is to predict an \emph{answer} $\answer \in \{ \yes,\no, \irrelevant \} \cup W^*$ that specifies whether the answer to the input question, in the context of the rule text and the previous follow-up question dialog, \MaxMargin{Do we also need scenario here?} is either \yes, \no, \irrelevant{} or another follow-up question in $W^*$. Here \irrelevant{} is the target answer whenever a rule text is not related to the question $\question$. 

\section{Annotation Protocol}

\begin{figure*}
\centering
\includegraphics[width=0.95\textwidth, angle=0]{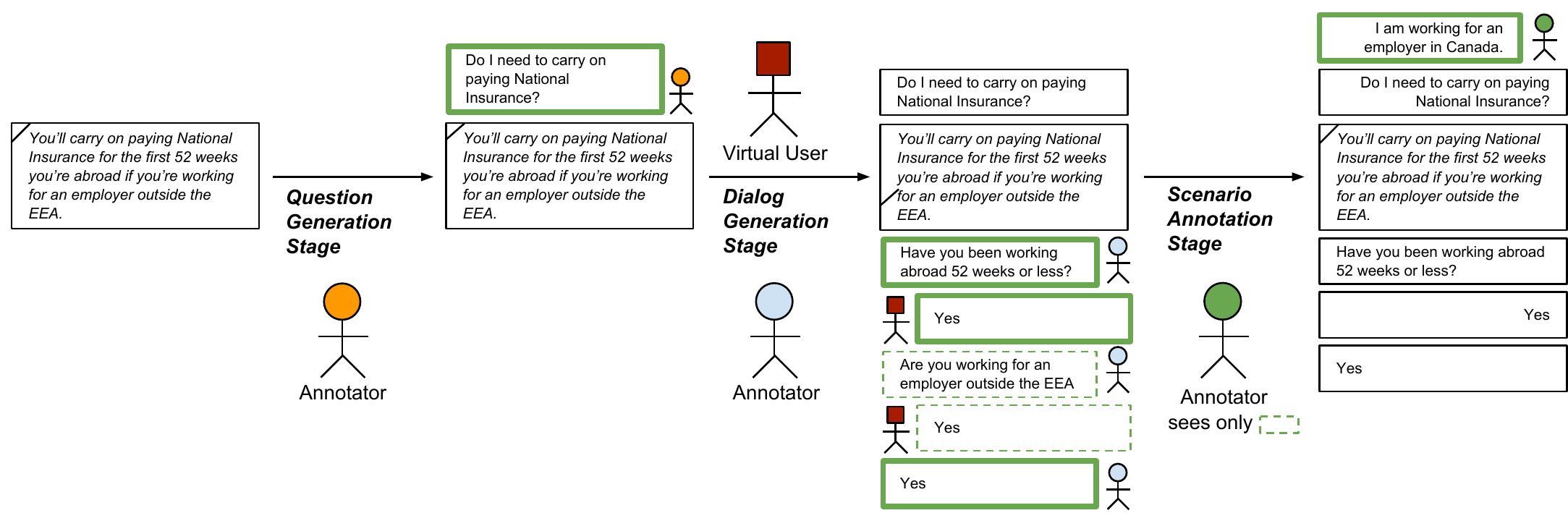}
\caption{\label{pipeline} The different stages of the annotation process (excluding the rule text extraction stage). First a human annotator generates an underspecified input question (question generation). Then, a virtual user and a human annotator collaborate to produce a dialog of follow-up questions and answers (dialog generation). Finally, a scenario is generated from parts of the dialog, and these parts are omitted in the final result.   }
\end{figure*}
% Source Document here: https://docs.google.com/drawings/d/1Go3cFXADLjiJHZoUs8IvyclmWG160KIsSegGPiHkeEI/edit

% \MaxMargin{We've said this before without really substantiating the claim} It is difficult for annotators to produce dialogs like the one in \Cref{example} from scratch. We hence develop a protocol to create annotations incrementally. 	
Our annotation protocol is depicted in \Cref{pipeline} and has four high-level stages: Rule Text Extraction, Question Generation, Dialog Generation and Scenario Annotation. We present these stages below, together with discussion of our quality-assurance mechanisms and method to generate negative data. For more details, such as annotation interfaces, we refer the reader to Appendix~\ref{appdx_annotation_interface}.  

% \MaxMargin{Maybe we should present this as a list of stages? Feels a bit hard to follow in text format}
% In the first stage~(omitted from the figure), a paragraph of text is chosen from a collection of documents. 
% In the second stage, an annotator produces an \emph{underspecified} question designed to require follow-up questions to be answered.
% In stage three, another annotator collaborates with a \emph{virtual user} to produce a dialog of follow-up questions given the input question and scenario. 
% In this dialog, we pretend that the virtual user is asking the initial question, and the human annotator is tasked to ask the right follow-up questions until they can infer an answer.
% In each turn, the virtual user randomly answers the annotator's follow-up question with \yes{} or \no{}. 
% In the final stage, a subset of the follow-up questions and answers is shown to a third annotator who in turn produces a scenario consistent with this subset. The final dialog contains this scenario, the initial question, the rule text, and the part of the dialog that was \emph{not} shown to the final annotator. 

\subsection{Rule Text Extraction Stage}
First, we identify the source documents that contain the rules we would like to annotate. Source documents can be found in Appendix~\ref{appdx_source}. 
% This can include  
% We scrape rule-based documents from government websites which permit the reproduction of this content for commercial and research purposes. Document sources included the UK, US and Australian government websites covering a variety of topics such as immigration regulations, benefit eligibility, tax credits, and livestock exportation requirements among others.  
We then convert each document to a set of \emph{rule texts} using a heuristic which identifies and groups paragraphs and bulleted lists.
% Consequently, we prepend the last encountered heading or sub-heading.
To preserve readability during the annotation, we also split by a maximum rule text length and a maximum number of bullets.

\subsection{Question Generation Stage} 
For each rule text we ask annotators to come up with an input question. Annotators are instructed to ask questions that cannot be answered directly but instead require follow-up questions. This means that the question should a) match the topic of the support rule text, and b) be underspecified. At present, this part of the annotation is done by expert annotators, but in future work we plan to crowd-source this step as well.%\footnote{Note that a single input question can lead to a large number of utterances because we can use human annotators to augment the question with a scenario in stage \itodo{}}. 

% We also develop an annotation interface in preparation for the eventual scaling-up of this dataset. 
 
\subsection{Dialog Generation Stage} 
In this stage, we view human annotators as assistants that help users reach the answer to the input question. 
Because the question was designed to be broad and to omit important information, human annotators will have to ask for this information using the rule text to figure out which question to ask. The follow-up question is then sent to a \emph{virtual user}, \ie{}, a program that simply generates a random \yes{} or \no{} answer. If the input question can be answered with this new information, the annotator should enter the respective answer. If not, the annotator should provide the next follow-up question and the process is repeated. 

% Note that the annotator in each turn can, and often will, change. This situation arises when we want to re-use a partial dialog, and  change the last answer of the virtual user. At this point, a second annotator can continue the conversation. 

%\subsubsection{Dialog Tree Coverage}
\begin{figure}
\centering
\includegraphics[width=0.40\textwidth, angle=0]{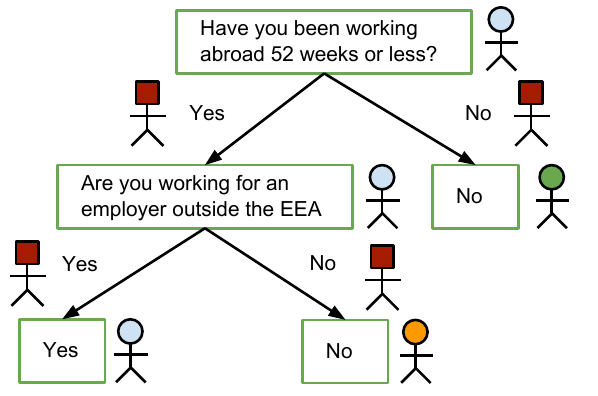}
\caption{\label{tree} We use different annotators (indicated by different colors) to create the complete dialog tree. }
\end{figure}
When the virtual user is providing random \yes{} and \no{} answers in the dialog generation stage, we are traversing a specific branch of a decision tree. We want the corpus to reflect all possible dialogs for each question and rule text. Hence, we ask annotators to label additional branches. For example, if the first annotator received a \yes{} as the answer to the second follow-up question in Figure~\ref{tree}, the second annotator (orange) receives a \no{}.

\subsection{Scenario Annotation Stage}\label{scenario-annotation}
In the final stage, we choose parts of the dialogs created in the previous stage and present this to an annotator. For example, the annotator sees ``Are you working or preparing for work?'' and \no. They are then asked to write a scenario that is consistent with this dialog such as ``I am currently out of work after being laid off from my last job, but am not able to look for any yet.''. The number of questions and answers that the annotator is presented with for generating a scenario can vary from one to the full length of a dialog. Users are encouraged to paraphrase the questions and not to use many words from the dialog.

In an attempt to make these scenarios closer to the real-world situations where a user may provide a lot of unnecessary information to an operator, not only do we present users with one or more questions and answers from a specific dialog but also with one question from a random dialog. The annotators are asked to come up with a scenario that fits all the questions and answers. 

Finally, a dialog is produced by combining the scenario with the input question and rule text from the previous stages. In addition, all dialog utterances that were \emph{not} shown to the final annotator are included as well as they complement the information in the scenario. Given a dialog of this form, we can create utterances that are described in Section~\ref{section_task}. 

As a result of this stage of annotation, we create a corpus of scenarios and questions where the correct answers (\yes{}, \no{} or \irrelevant{}) to questions can be derived from the related scenarios. This corpus and its challenges will be discussed in Section~\ref{scenario_interpretation}. \todo{Is it only correct answers or also necessary follow up questions}

\subsection{Negative Examples}\label{section_neg}
To facilitate the future application of the models to large-scale rule-based documents instead of rule text, we deem it to be imperative for the data to contain negative examples of both questions and scenarios.

% \paragraph{Negative Questions}
We define a \emph{negative question} as a question that is not relevant to the rule text. In this case, we expect models to produce the answer \irrelevant{}. For a given rule text and question pair, a negative example is generated by sampling a random question from the set of all possible questions, excluding the question itself and questions sourced from the same document using a methodology similar to the work of \newcite{levy2017zero}.

%\paragraph{Negative Scenarios}
\MaxMargin{I would rephrase, time permitting. Not sure how. Can also be removed.} The data created so far is biased in the sense that when a scenario is given, at least one of the follow-up questions in a dialog can be answered. In practice, we expect users to also provide background scenarios that are completely irrelevant to the input question. Therefore, we sample a \emph{negative scenario} for each input question and rule text pair, $(q,r)$ in our data. We uniformly sample from the scenarios created in \Cref{scenario-annotation} for all question and rule text pairs $(q',r')$ unequal to $(q,r)$. \MaxMargin{Should we have the negative data validation results here? They should still be applicable} For more details, we point the reader to Appendix~\ref{appdx_negative_data}.

\subsection{Quality Control}\label{validation}
We employ a range of quality control measures throughout the process. In particular, we:
\begin{enumerate}[nosep]
\item Re-annotate pre-terminal nodes in the dialog trees if they have identical \yes{} and \no{} branches. 
\item Ask annotators to validate the previous dialog in case previous utterances where created by different annotators. 
\item Assess a sample of annotations for each annotator and keep only those annotators with quality scores higher than a certain threshold. 
\item We require annotators to pass a qualification test before selecting them for our tasks. We also require high approval rates and restrict location to the UK, US, or Canada. 
\end{enumerate}
Further details are provided in Appendix~\ref{appdx_quality_control}.

\subsection{Cost, Duration and Scalability} 
The cost of different stages of annotation is as follows. An annotator was paid \$0.15 for an initial question (948 questions), \$0.11 for a dialog part (3000 dialog parts) and \$0.20 for a scenario (6,600 scenarios). It takes in total 2 weeks to complete the annotation process. 
Considering that all the annotation stages can be done through crowdsourcing and in a relatively short time period and at a reasonable cost using established validation procedures, the dataset can be scaled up without major bottlenecks or an impact on the quality.

\section{\dsacr{}}
In this section, we present the \textit{\dsfull{} (\dsacr{})} dataset.\footnote{The dataset and its Codalab challenge can be found at \url{https://sharc-data.github.io}.} 

% \subsection{Source Data}
% We use $264$ unique sources from $10$ unique domains, e.g. \url{www.uscis.gov} and \url{www.benefits.gov}. For transparency, the source URLs are included in the corpus for each dialog utterance. xx

\subsection{Dataset Size and Quality}

The dataset is built up from of 948 distinct snippets of rule text. Each has an input question and a ``dialog tree''. At each step in the dialog, there is a followup question posed and the tree branches depending on the answer to the followup question (yes/no). The \dsacr{} dataset is comprised of all individual ``utterances'' from every tree, i.e. every possible point/node in any dialog tree. There are 6058 of these utterances. In addition, there are 6637 scenarios that provide more information, allowing some questions in the dialog tree to be ``skipped'' as the answers can be inferred from the scenario. Scenarios therefore modify the dialog trees, which creates new trees. When combined with scenarios and negative sampled scenarios, the total number of distinct utterances became 37087. As a final step, utterances were removed where the scenario referred to a portion of the dialog tree that was unreachable for that utterance, leaving a final dataset size of 32436 utterances.\footnote{One may argue that the the size of the dataset is not sufficient for training end-to-end neural models. While we believe that the availability of large datasets such as SNLI or SQuAD has helped drive the state-of-the-art forward on related tasks, relying solely on large datasets to push the boundaries of AI cannot be as practical as developing better models for incorporating common sense and external knowledge which we believe \dsacr{} is a good test-bed for. Furthermore, the proposed annotation protocol and evaluation procedure can be used to reliably extend the dataset or create datasets for new domains.}

We break these into train, development and test sets such that each dataset contains approximately the same proportion of sources from each domain, targeting a 70\%/10\%/20\% split. 

% \subsection{Dataset Quality Metrics}
% We can assess the quality of \dsacr{} using the validation methods proposed in section~\ref{validation}. We find that \itodo{}.

% \subsubsection{Dialog Generation}
To evaluate the quality of dialog generation HITs, we sample a subset of 200 rule texts and questions and allow each HIT to be annotated by three distinct workers. In terms of deciding whether the answer is a \yes{}, \no{} or some follow-up question, the three annotators reach an answer agreement of $72.3\%$. \SebMargin{Is this yes/more? This should be linked to previous section} We also calculate Cohen's Kappa, a measure designed for situations with two annotators. We randomly select two out of the three annotations and compute the unweighted kappa values, repeated for 100 times and averaged to give a value of $0.82$.

% \paragraph{Joint Probability of Agreement}
% \SebMargin{can this be shortened? Can some of this be moved up?}
% This metric measures the agreement rate between annotators on whether they respond with a \yes{}, \no{}, or a \followupquestion{}, to a given support document, a question and partial or complete dialogue history. We find that the greatest proportion of disagreements occurred when the worker predicted a \yes{} response in a situation that the majority determined required a \followupquestion{}. We randomly select our 200 samples from all previously generated data excluding initial utterances. This is because our annotation protocol is designed with the intent of obtaining a \followupquestion{} response in the initial setting and failing to recognise and account for this would artificially inflate our agreement scores. We register a tri-annotator joint agreement probability of $72.3\%$.

% \paragraph{Cohen's Kappa} 
% Cohen's Kappa is designed for situations with two annotators. We therefore randomly select two out of the three annotations and compute the unweighted kappa values, repeated for 100 times and averaged to give a value of $0.82$.

% \paragraph{Fleiss' Kappa} This statistic, originally presented as a generalisation of Cohen's Kappa \cite{fleiss1971measuring}, adapts this and Scott's Pi to the multiple-rater setting. We obtain a Fleiss' Kappa score of $0.71$.

%\paragraph{BLEU Score}
\SebMargin{this should be shortened}
\MaxMargin{Should we make it a bit more explicit e.g. BLEU Score between Annotations or something a bit better explained?}
The above metrics measure whether annotators agree in terms of deciding between \yes{}, \no{} or some follow-up question, but not whether the follow-up questions are equivalent. To approximate this, we calculate BLEU scores between pairs of annotators when they both predict follow-up questions.
% Given the support document, a question and a history, we measure the similarity between the follow-up questions of annotators. This shows the degree of agreement in which annotators think a clarification question is necessary to answer the question. In many cases, more than one follow-up questions can be asked and the annotators can choose different orders to ask these questions. To take this into account, for each follow-up question of the first annotator, we calculate the BLEU score with all the follow-up questions of the second annotator and take the maximum value. We do this vice versa and take the average as the BLEU score. Since the BLEU score is calculated between two annotators, we randomly choose two annotations amongst the available three for each utterance. We run this process several times and average the scores. 
Generally, we find high agreement: Annotators reach average BLEU scores of $0.71$, $0.63$, $0.58$ and $0.58$ for maximum orders of $1$, $2$, $3$ and $4$ respectively.

%\paragraph{Human Performance} 
To get an indication of human performance on the sub-task of classifying whether a response should be a \yes{}, \no{} or \followupquestion{}, we use a similar methodology to \cite{Rajpurkar2016_SQUAD} by considering the second answer to each question as the human prediction and taking the majority vote as ground truth. The resulting human accuracy is $93.9\%$.

%\subsubsection{Scenario Generation}
\MaxMargin{I think this whole section can go. Annotator agreement in this context gives no indication of the quality of the data as it is not agreement between annotators contributing to the actual dataset. The bit of value in this paragraph is that approval rate increased from 38\% without qualification to 93\% with (I think it was something like 93 because we annotated less than 100 - don't know where the data is), which is relevant to section 3.6(4) and should be moved there.}
To evaluate the quality of the scenarios, we sample 100 scenarios randomly and ask two expert annotators to validate them. We perform validation for two cases: 1) scenarios generated by turkers who did not attempt the qualification test and were not filtered by our validation process, 2) scenarios that are generated by turkers who have passed the qualification test and validation process. In the second case, annotators approved an average of 89 of the scenarios whereas in the first case, they only approved an average of 38. This shows that the qualification test and the validation process improved the quality of the generated scenarios by more than double. In both cases, the annotators agreed on the validity of 91-92 of the scenarios. For further details on dataset quality, the reader is referred to Appendix~\ref{appdx_quality_control}.
% \footnote{In 8\%-9\% of cases, annotators did not agree whether scenarios imply all the provided dialog parts. This shows one of the challenges of a machine reading system that needs to infer information from the provided scenarios.}

\subsection{Challenges}
\MaxMargin{This paragraph is redundant}
We analyse the challenges involved in solving conversational machine reading in \dsacr{}. We divide these into two parts: challenges that arise when interpreting rules, and challenges that arise when interpreting scenarios. 

\subsubsection{Interpreting Rules}
When no scenarios are available, the task reduces to \MaxMargin{I find these a), b) c) things hard to read. I think they should either turn into standard text or become lists.} a) identifying the follow-up questions within the rule text, b) understanding whether a follow-up question has already been answered in the history, and c) determining the logical structure of the rule (\eg{} disjunction vs. conjunction vs. conjunction of disjunctions) \MaxMargin{what about disjunction of conjunctions? I'd say: e.g. conjunctions, disjunctions and complex combinations}. 

To illustrate the challenges that these sub-tasks involve, we manually categorise a random sample of 100 $\left(\question_i, \support_i\right)$ pairs. We identify 9 phenomena of interest, and estimate their frequency within the corpus. Here we briefly highlight some categories of interest, but full details, including examples, can be found in Appendix~\ref{appdx_interpreting_rules}. 
% in Table~\ref{table_rule_categories}, and plotted in figure~\ref{challenges_followup_answers_reasoning}. 

A large fraction of problems involve the identification of at least two conditions, and approximately 41\% and 27\% of the cases involve logical disjunctions and conjunctions respectively. These can appear in linguistic coordination structures as well as bullet points. Often, differentiating between conjunctions and disjunctions is easy when considering bullets---key phrases such as ``if all of the following hold'' can give this away. However, in 13\% of the cases, no such cues are given and we have to rely on language understanding to differentiate. For example:\vspace{0.2cm}      
{%\small
\resizebox{\columnwidth}{!}{
\begin{tikzpicture}
\node[draw, rectangle, text width=7.55cm] (q) {\textbf{Q}: Do I qualify for Statutory Maternity Leave?};
\node[draw, rectangle, fill=red!5, text width=7.55cm, below = -0.05cm of q] (r) {\textbf{R}: You qualify for Statutory Maternity Leave if\\
\hspace{0.5cm}- you're an employee not a ``worker''\\
\hspace{0.5cm}- you give your employer the correct notice};
\end{tikzpicture}
}
}

% In this example, whether to combine the two bullet points as a disjunction or conjunction is not clear from syntactic clues, and a model that relies on learning shallow syntactic patterns cannot successfully answer. In order to arrive at the correct answer ( a conjunction ), models must be able to infer the relationship between employees and employers that does not exist between workers and employers, and thus conclude that the bullet points must be conjunctive.

\subsubsection{Interpreting Scenarios}\label{scenario_interpretation}
Scenario interpretation can be considered as a multi-sentence entailment task. Given a scenario (premise) of (usually) several sentences, and a question (hypothesis), a system should output \yes{} (\entailment{}), \no{} (\contradiction{}) or \irrelevant{} (\neutral{}). In this context, \irrelevant{} indicates that the answer to the question cannot be inferred from the scenario. 

%While many entailment corpora such as SNLI~\cite{bowman2015large} provide single sentences as the premise and the hypothesis, the majority of our scenarios contain more than one sentence and carry several pieces of information. 
% This can make the entailment task more challenging. 

Different types of reasoning are required to interpret the scenarios. Examples include numerical reasoning, temporal reasoning and implication (common sense and external knowledge). We manually label 100 scenarios with the type of reasoning required to answer their questions. Table~\ref{table_scenario_categories} shows examples of different types of reasoning and their percentages. Note that these percentages do not add up to 100\% as interpreting a scenario may require more than one type of reasoning.

\begin{table*}[t]
\centering
\small
\begin{tabular}{p{11mm} p{59mm} p{68mm} p{5mm}}
\toprule
\multicolumn{1}{ l }{\textbf{Category}} & \multicolumn{1}{ c }{\textbf{Questions}} & \multicolumn{1}{ c }{\textbf{Scenario}} & \textbf{\%}\\
% \hline
\midrule
Explicit & Has your wife reached state pension age? \underline{Yes} & My wife just recently reached the age for state pension & 25\%\\
\addlinespace
%\midrule
Temporal & Did you own it before April 1982? \underline{Yes} & I purchased the property on June 5, 1980. & 10\%\\
\addlinespace
%\midrule
Geographic & Do you normally live in the UK? \underline{No} & I'm a resident of Germany. & 7\%\\
\addlinespace
%\midrule
Numeric & Do you work less than 24 hours a week between you? \underline{No} & My wife and I work long hours and get between 90 - 110 hours per week between the two of us. & 12\%\\
\addlinespace
%\midrule
Paraphrase & Are you working or preparing for work? \underline{No} & I am currently out of work after being laid off from my last job, but am not able to look for any yet. & 19\%\\
\addlinespace
%\midrule
Implication & Are you the baby's father? \underline{No}  & My girlfriend is having a baby by her ex. & 51\%\\
\bottomrule
\end{tabular}
\caption{Types of reasoning and their proportions in the dataset based on 100 samples. Implication includes reasoning beyond what is explicitly stated in the text, including common sense reasoning and external knowledge.}
\label{table_scenario_categories}
\end{table*}

% \subsection{Cost}
% For the dialog generation stage, we run a total of $4040$ HITs (including those not validated), each HIT taking a mean time of 37.1s to complete and a cost of $\$0.12$. For the scenario generation stage, we run a total of approximately $9656$ HITs, with each HIT taking an average of $200$ seconds. Each scenario generation HIT costs either $\$0.2$ or $\$0.3$, depending on the number of questions each scenario has to cover, \MaxMargin{Don't think we should quote total cost} with a total cost of $\tilde{} 2,481$.

\section{Experiments}
To assess the difficulty of \dsacr{} as a machine learning problem, we investigate a set of baseline approaches on the end-to-end task as well as the important sub-tasks we identified. The baselines are chosen to assess and demonstrate both feasibility and difficulty of the tasks.
%We start with these subtasks: yes/no/more classification, follow-up question generation and scenario interpretation. Then we investigate performance on the end-to-end task. 

% We also ask whether a CMR system, even without scenario interpretation, can be useful for human users. To this end we conduct a user study that \itodo{}...

\paragraph{Metrics}
For all following classification tasks, we use micro- and macro- averaged accuracies. \MaxMargin{I believe this is incorrect. We compute BLEU between gold follow-up question $y_i$ and predicted $y^hat_i$ no?} For the follow-up generation task, we compute the BLEU scores at orders 1, 2, 3 and 4 computed  between the gold follow-up questions, $\answer_i$ and follow-up question $\hat \answer_i = \word_{\hat y_i,1},\word_{\hat y_i,2} \ldots \word_{\hat y_i,n} $ for all utterances $i$ in the evaluation dataset.

\subsection{Classification (excluding Scenarios)}
\MikeMargin{Perhaps this explanation should happen sooner?} On each turn, a CMR system needs to decide, either explicitly or implicitly, whether the answer is \yes{} or \no{}, whether the question is not relevant to the rule text (\irrelevant{}), or whether a follow-up question is necessary---an outcome we label as \more{}. In the following experiments, we will test whether one can learn to make this decision using the \dsacr{} training data.

When a non-empty scenario is given, this task also requires an understanding of how scenarios answer follow-up questions. In order to focus on the challenges of rule interpretation, here we only consider empty scenarios. 

Formally, for an utterance $x = (\question, \support, \history, \scenario)$, we require models to predict an answer $\answer$ where $\answer \in \{\yes, \no, \irrelevant, \more \} $. Since we consider only the classification task without scenario influence, we consider the subset of utterances such that $\scenario = NULL$. This data subset consists of $4026$ train, $431$ dev and $1601$ test utterances.

% \subsubsection{Metrics}
% We look at the micro and macro- averaged accuracies over the four labels.
\subsubsect{Baselines}
We evaluate various baselines including random, a surface logistic regression applied to a TFIDF representation of the rule text, question and history, a rule-based heuristic which makes predictions depending on the number of overlapping words between the rule text and question, detecting conjunctive or disjunctive rules, detecting negative mismatch between the rule text and the question and what the answer to the last follow-up history was, a feature-engineered Random Forest and a Convolutional Neural Network applied to the tokenised inputs of the concatenated rule text, question and history.

\subsubsect{Results}
We find that, for this classification sub-task, Random Forest slightly outperforms the heuristic. All learnt models considerably outperform the random and majority baselines.

\Max{I would have these results tables positioned [h] to appear in place so that the results flow on from the text. Like this, I find looking at the results rather confusing (especially since we have so many different types)}

\begin{table}
\small
\centering
\begin{tabular}{ l c c }
\toprule
Model & Micro Acc. & Macro Acc.\\
\midrule
Random & $0.254$ & $0.250$\\
% Majority & $0.314$ & $0.250$\\
Surface LR & $0.555$ & $0.511$\\
Heuristic & $0.791$ & $0.779$\\
Random Forest & $\mathbf{0.808}$ & $\mathbf{0.797}$\\
\addlinespace
CNN & $0.677$ & $0.681$\\
\bottomrule
\end{tabular}
\label{classification_results_table}
\caption{Selected Results of the baseline models on the classification sub-task.}
\end{table}

\subsection{Follow-up Question Generation without Scenarios}

When the target utterance is a follow-up question, we still have to determine what that follow-up question is. For an utterance $x = (\question, \support, \history, \scenario)$, we require models to predict an answer $\answer$ where $\answer$ is the next follow-up question, $\answer = \word_{y,1},\word_{y,2} \ldots \word_{y,n} = f_{m+1}$ if $x$ has history of length $m$. We therefore consider the subset of utterances such that $\scenario = NULL$ and $\answer \not \in \{\yes, \no, \irrelevant \} $. This data subset consists of 1071 train, 112 dev and 424 test utterances.

% \subsubsection{Metrics}

% We evaluate systems for this task using BLEU scores. Specifically, we compute the BLEU scores at maximum orders 1, 2, 3 and 4 computed with between the gold follow-up questions, $\answer_i$ and follow-up question $\hat \answer_i = \word_{\hat y_i,1},\word_{\hat y_i,2} \ldots \word_{\hat y_i,n} $ for all instances $i$ in the evaluation dataset.

\subsubsect{Baselines}

We first consider several simple baselines to explore the relationship between our evaluation metric and the task. As annotators are encouraged to re-use the words from rule text when generating follow-up questions, a baseline that simply returns the final sentence of the rule text performs surprisingly well. We also implement a rule-based model that uses several heuristics.

If framed as a seq2seq task, a modified CopyNet is most promising \cite{gu2016copy}. We also experiment with span extraction/sequence-tagging approaches to identify relevant spans from the rule text that correspond to the next follow-up questions. We find that Bidirectional Attention Flow \cite{Seo2016BidAF} performed well.\footnote{\label{allennote}We use AllenNLP implementations of BiDAF \& DAM}\addtocounter{footnote}{-1}\addtocounter{Hfootnote}{-1} Further implementation details can be found in Appendix~\ref{appdx_question_gen}.

\subsubsect{Results}
Our results, shown in Table \ref{followup_results_table} indicate that systems that return contiguous spans from the rule text perform better according to our BLEU metric. We speculate that the logical forms in the data are challenging for existing models to extract and manipulate, which may suggest why the explicit rule-based system performed best.
We further note that only the rule-based and NMT-Copy models are capable of generating genuine questions rather than spans or sentences.

\begin{table}
\small
\centering
\begin{tabular}{ l c c c c }
\toprule
Model & BLEU-1 & BLEU-2 & BLEU-3 & BLEU-4 \\
\midrule
%Rand Sent.& $0.302$ & $0.228$ & $0.197$ & $0.179$\\
First Sent.& $0.221$ & $0.144$ & $0.119	$ & $0.106$ \\
%Last Sent.& $0.314$ & $0.247$ & $0.217$ & $0.197$ \\
%Surface LR & $0.293$ & $0.233$ & $0.205$ & $0.186$ \\
\addlinespace
NMT-Copy & $0.339$ & $0.206$ & $0.139$ & $0.102$\\
%Seq Tag & $0.212$&$0.151$&$0.126$&$0.110$\\
BiDAF & $0.450$&$0.375$&$0.338$&$0.312$\\
Rule-based & $\mathbf{0.533}$&$\mathbf{0.437}$&$\mathbf{0.379}$&$\mathbf{0.344}$\\
\bottomrule
\end{tabular}
\caption{Selected Results of the baseline models on follow-up question generation.}
\label{followup_results_table}
\end{table}

\subsection{Scenario Interpretation}
\MaxMargin{I would quantify ``many'' or exclude} Many utterances require the interpretation of the scenario associated with a question. If the scenario is understood, certain follow-up questions can be \MaxMargin{skipped -> inferred? or something better? consider rewording} skipped because they are answered within the scenario. 
% For example, given the scenario ``I am currently out of work after being laid off from my last job, but am not able to look for any yet.'' and the question ``Are you working or preparing for work?'' the answer should be \no{} or ``contradiction''.
In this section, we investigate how \MaxMargin{I'm not sure we're really answering the research question ``How difficult is scenario interpretation?'' based on the models we've trained. Maybe we've provided insight into some approaches one might consider taking?} difficult scenario interpretation is by training models to answer follow-up questions based on scenarios. 

\subsubsect{Baselines}
We use a random baseline and also implement a surface logistic regression applied to a TFIDF representation of the combined scenario and the question. For neural models, we use Decomposed Attention Model (DAM)~\citep{parikh2016decomposable} trained on each the SNLI and \dsacr{} corpora using ELMO embeddings~\cite{peters2018deep}.\footnotemark

\subsubsect{Results}
Table~\ref{table_results_entailment} shows the result of our baseline models on the entailment corpus of \MikeMargin{the?} \dsacr{} test set. Results show poor performance especially for the macro accuracy metric of both simple baselines and neural state-of-the-art entailment models. This performance highlights the challenges that the scenario interpretation task of \dsacr{} presents, many of which are discussed in Section~\ref{scenario_interpretation}.
\begin{table}
\centering
\small
\begin{tabular}{l c c}
\toprule
Model & Micro Acc. & Macro Acc.\\
\midrule
Random & $0.330$ & $0.326$\\
Surface LR & $\mathbf{0.682}$ & $0.333$ \\
\addlinespace
DAM (SNLI) & $0.479$ & $\mathbf{0.362}$ \\
DAM (\dsacr{}) & $0.492$ & $0.322$ \\
\bottomrule
\end{tabular}
\caption{Results of entailment models on \dsacr{}.}
\label{table_results_entailment}
\end{table}

\subsection{Conversational Machine Reading}
The CMR task requires all of the above abilities. To understand its core challenges, we compare baselines that are trained end-to-end vs. baselines that reuse solutions for the above subtasks. 
% \subsubsection{Metrics}
% We consider the micro and macro- averaged accuracies over all utterances for the four labels \yes{}, \no{}, \more{} and \irrelevant{} where the evaluator considers any generated follow-up question as \more{}. We also compute the BLEU scores for any predictions where the ground truth label was some follow-up question.
\subsubsect{Baselines}
We present a Combined Model (CM) which is a pipeline of the best performing Random Forest classification model, rule-based follow-up question generation model and Surface LR entailment model. We first run the classification model to predict \yes{}, \no{}, \more{} or \irrelevant{}. If \more{} is predicted, the Follow-up Question Generation model is used to produce a follow-up question, $f_1$. The rule text and produced follow-up question are then passed as inputs to the Scenario Interpretation model. If the output of this is \irrelevant{}, then the CM predicts $f_1$, otherwise, these steps are repeated recursively until the classification model no longer predicts \more{} or the entailment model predicts \irrelevant{}, in which case the model produces a final answer. We also investigate an extension of the NMT-copy model on the end-to-end task. Input sequences are encoded as a concatenation of the rule text, question, scenario and history. The model consists of a shared encoder LSTM, a 4-class classification head with attention, and a decoder GRU to generate followup questions. The model was trained by alternating training the classifier via standard softmax-cross entropy loss and the followup generator via seq2seq. At test time, the input is first classified, and if the predicted class is \more{}, the follow-up generator is used to generate a followup question, $f_1$. A simpler model without the separate classification head failed to produce predictive results.

\subsubsect{Results}
\MaxMargin{Just wrote this. Please review and ensure it fits with overall narrative} We find that the combined model outperforms the neural end-to-end model on the CMR task, however, the fact that the neural model has learned to classify better than random and also predict follow-up questions is encouraging for designing more sophisticated neural models for this task. %motivates the development or adaptation of neural models better suited to this task.
\MikeMargin{Either needs some text referencing the table or force that table to display here}
\begin{table}
\small
\centering
\setlength{\tabcolsep}{5pt}
\begin{tabular}{ l  c  c  c  c }
\toprule
\bf Model & \bf Micro Acc & \bf Macro Acc & \bf BLEU-1 & \bf BLEU-4 \\
\midrule
CM & $0.619$ & $0.689$ & $0.544$ & $0.344$\\
NMT & $0.448$ & $0.428$ & $0.340$ & $0.078$\\
\bottomrule
\end{tabular}
\caption{Results of the models on the CMR task.}
\label{combined_table}
\vskip -5mm
\end{table}

\subsubsect{User Study}
In order to evaluate the utility of conversational machine reading, we run a user study that compares CMR to when such an agent is not available, i.e. the user has to read the rule text and determine themselves the answer to the question. % is \quoteYes{} or \quoteNo{}.
On the other hand, with the agent, the user does not read the rule text, instead only responds to follow-up questions. % with a \quoteYes{} or \quoteNo{}, based on the scenario text and world knowledge.
Our results show that users using the conversational agent reach conclusions $>2$ times faster than ones that are not, but more importantly, they are also much more accurate ($93\%$ as compared to $68\%$). %, while the users that are manually read and reason are often incorrect (accuracy of $68\%$ for a binary classification task).
Details of the experiments and the results are included in Appendix~\ref{app:e2e}.

\section{Related Work}
This work relates to several areas of active research. 
\paragraph{Machine Reading}
In our task, systems answer questions about units of texts. In this sense, it is most related to work in Machine Reading~\cite{Rajpurkar2016_SQUAD,Seo2016BidAF,Weissenborn2017fastQA}. The core difference lies in the conversational nature of our task: in traditional Machine Reading the questions can be answered right away; in our setting, clarification questions are often needed. The domain of text we consider is also different (regulatory vs Wikipedia, books, newswire). \Seb{Maybe add reasoning and Wikihop}     
\paragraph{Dialog}
% \SebMargin{cite some old school Dialog papers} 
The task we propose is, at its heart, about conducting a dialog~\cite{weizenbaum_elizacomputer_1966,serban_survey_2018,DBLP:journals/corr/BordesW16}. Within this scope, our work is closest to work in dialog-based QA where complex information needs are addressed using a series of questions. In this space, previous approaches have been looking primarily at QA dialogs about images~\cite{das_visual_2017} and knowledge graphs~\cite{saha_complex_2018,iyyer_search-based_2017}. In parallel to our work, both \newcite{choi_quac_2018} and \newcite{reddy_coqa:_2018} have to began to investigate QA dialogs with background text. Our work not only differs in the domain covered  (regulatory text vs wikipedia), but also in the fact that our task requires the interpretation of complex rules, application of background knowledge, and the formulation of free-form clarification questions. \newcite{rao_learning_2018} investigate how to generate clarification questions but this does not require the understanding of explicit natural language rules.

% The task we propose is, at its heart, about conducting a dialog. There has been a lot of recent work on dialog generation/management in general and end-to-end neural dialog systems in particular~\cite{Serban:2016:BED:3016387.3016435,DBLP:journals/corr/BordesW16}. So far, the developed systems lack the ability to use background text to answer questions, derive clarification questions based on the background text, or work on regulatory text. \Seb{Reviewers from ACL had some comments here, no?} \Max{Yep: Weakness argument 5: Insufficient contextualization with respect to other conversational agents work. While this emphasizes finding sequences of questions, we are really interested in observing user state based on the outcome of questions. This would fit in the standard POMDP-type formalization (e.g., Steve Young, Jason Williams, etc.) -- from this interpretation, we are mostly just producing utterances in the absence of a corpus (and doing it randomly). Basically, I think the related work section on dialogue is insufficient (and not even citing the most related work).}   

\paragraph{Rule Extraction From Text}
There is a long line of work in the automatic extraction of rules from text~\cite{SILVESTRO1988159,163670,Delisle94fromtext,Hassanpour11Framework,163670}. The work tackles a similar problem---interpretation of rules and regulatory text---but frames it as a text-to-structure task as opposed to end-to-end question-answering. For example, \newcite{Delisle94fromtext} maps text to horn clauses. This can be very effective, and good results are reported, but suffers from the general problem of such approaches: they require careful ontology building, layers of error-prone linguistic preprocessing, and are difficult for non-experts to create annotations for. 

\paragraph{Question Generation}
Our task involves the automatic generation of natural language questions. Previous work in question generation has focussed on producing questions for a given text, such that the questions can be answered using this text~\cite{Vanderwende08theimportance,olney12,Rus:2011:QGS:2187681.2187740}. \SebMargin{I have more citations here, but somehow the latex breaks when I add them. Must be an issue with the bibtex file}  
In our case, the questions to generate are \emph{derived} from the background text but cannot be answered by them. \newcite{Mostafazadeh2016GeneratingNQ} investigate how to generate natural follow-up questions based on the content of an image. Besides not working in a visual context, our task is also different because we see question generation as a sub-task of question answering.

% \paragraph{Annotating Semantics via QA}
% There has been recent work that started to cast tasks traditionally seen as text-to-structure into question answering tasks. For example, \newcite{he15Question} frames semantic role labeling as a question-answering task by
% using question-answer pairs to specify verbal
% arguments and the roles they play. Our work differs in the size of the inputs to annotate (paragraphs) and the phenomena we intend to capture (conditionals, negations, interpretation of coordination structures). 
% \begin{itemize}
% \item \cite{Chaudhri17Acquiring}---focuses on highly regular rules, not real natural language.
% \item \cite{Papanikolaou12Natural}---odd hack/system
% \item \cite{Hassanpour11Framework}---relies on ontologies, very specific application domain, hard to scale up. 
% \item \cite{Delannoy95}---not sure yet.
% \item \cite{Delisle94fromtext}---a lot of manual engineering (very useful related work section)
% \item \cite{163670}---can't access text, but sounds super relevant
% \item \cite{SILVESTRO1988159}---can't access, again very relevant.
% \end{itemize}
\section{Conclusion}
In this paper we present a new task as well as an annotation protocol, a dataset, and a set of baselines. The task is challenging and requires models to generate language, copy tokens, and make logical inferences. Through the use of an interactive and dialog-based annotation interface, we achieve good agreement rates at a low cost. \MaxMargin{Repeated} Initial baseline results suggest that substantial improvements are possible and require sophisticated integration of entailment-like reasoning and question generation.   %The task is challenging, but

\section*{Acknowledgements}
This work was supported by in part by an Allen Distinguished Investigator Award and in part by Allen Institute for Artificial Intelligence (AI2) award to UCI.
% ??

% \clearpage
\bibliography{riedel}
\bibliographystyle{acl_natbib_nourl}
\clearpage

% \makeatletter
% \setlength{\@fptop}{0pt}
% \makeatother
% \begin{table*}[ht!]
% \begin{center}
\twocolumn[
\centering
{\Large \emph{Supplementary Materials for EMNLP 2018 Paper:}}\\
\smallskip
{\Large Interpretation of Natural Language Rules in Conversational Machine Reading}\\
\bigskip
]
% \end{center}
% \end{table*}

%\clearpage
\appendix
\section{Annotation Interfaces}\label{appdx_annotation_interface}
\begin{figure}[h]
\includegraphics[width=0.5\textwidth, angle=0]{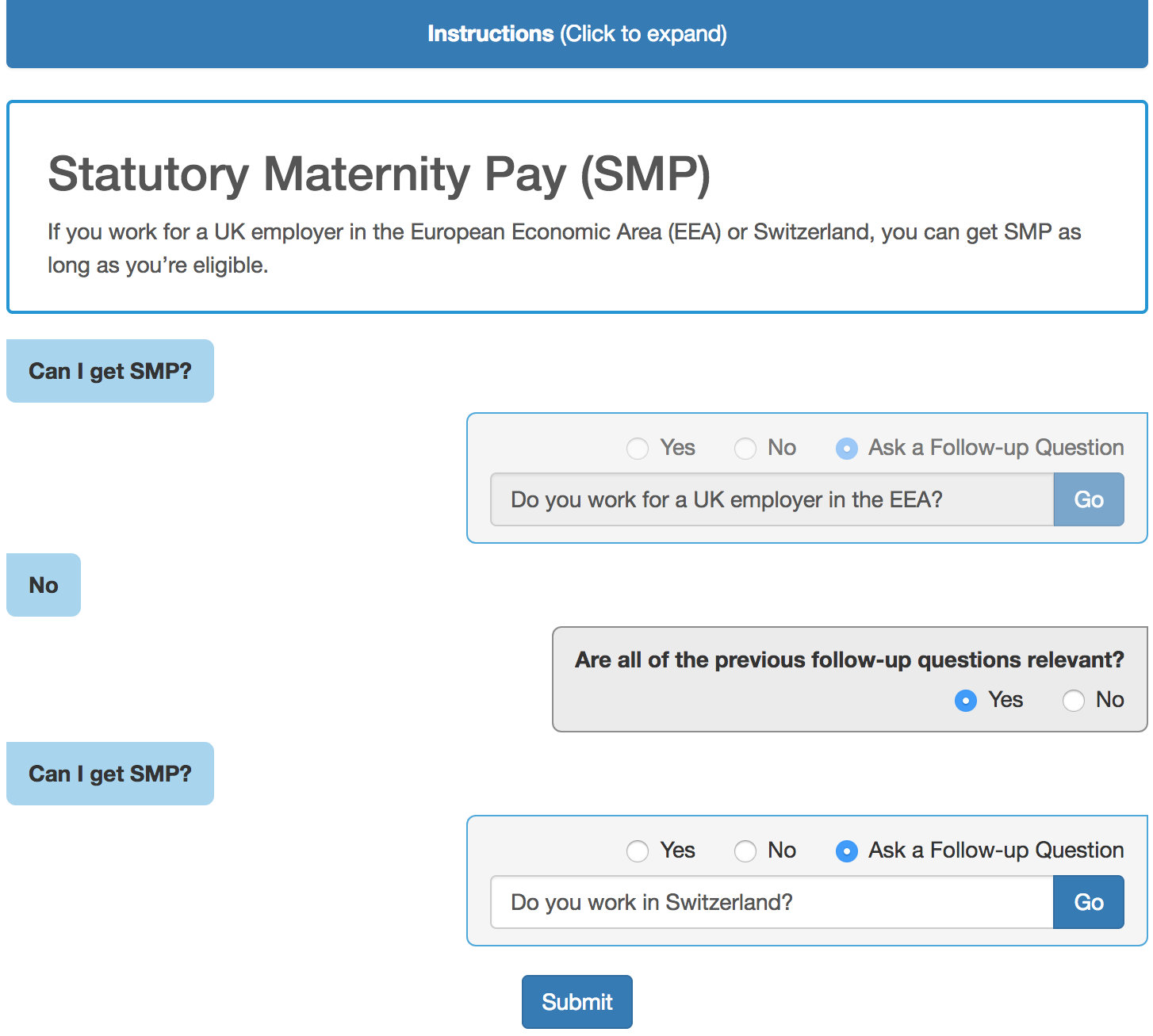}
\caption{\label{interface} The dialog-style web interface encourages workers to extract all the rule text-relevant evidence required to answer the initial question in the form of \yes{}/\no{} follow-up questions. }
\end{figure}

Figure~\ref{interface} shows the Mechanical-Turk interface we developed for the dialog generation stage. Note that the interface also contains a mechanism to validate previous utterances in case they have been generated by different annotators. 

\begin{figure}[h]
\includegraphics[width=0.45\textwidth]{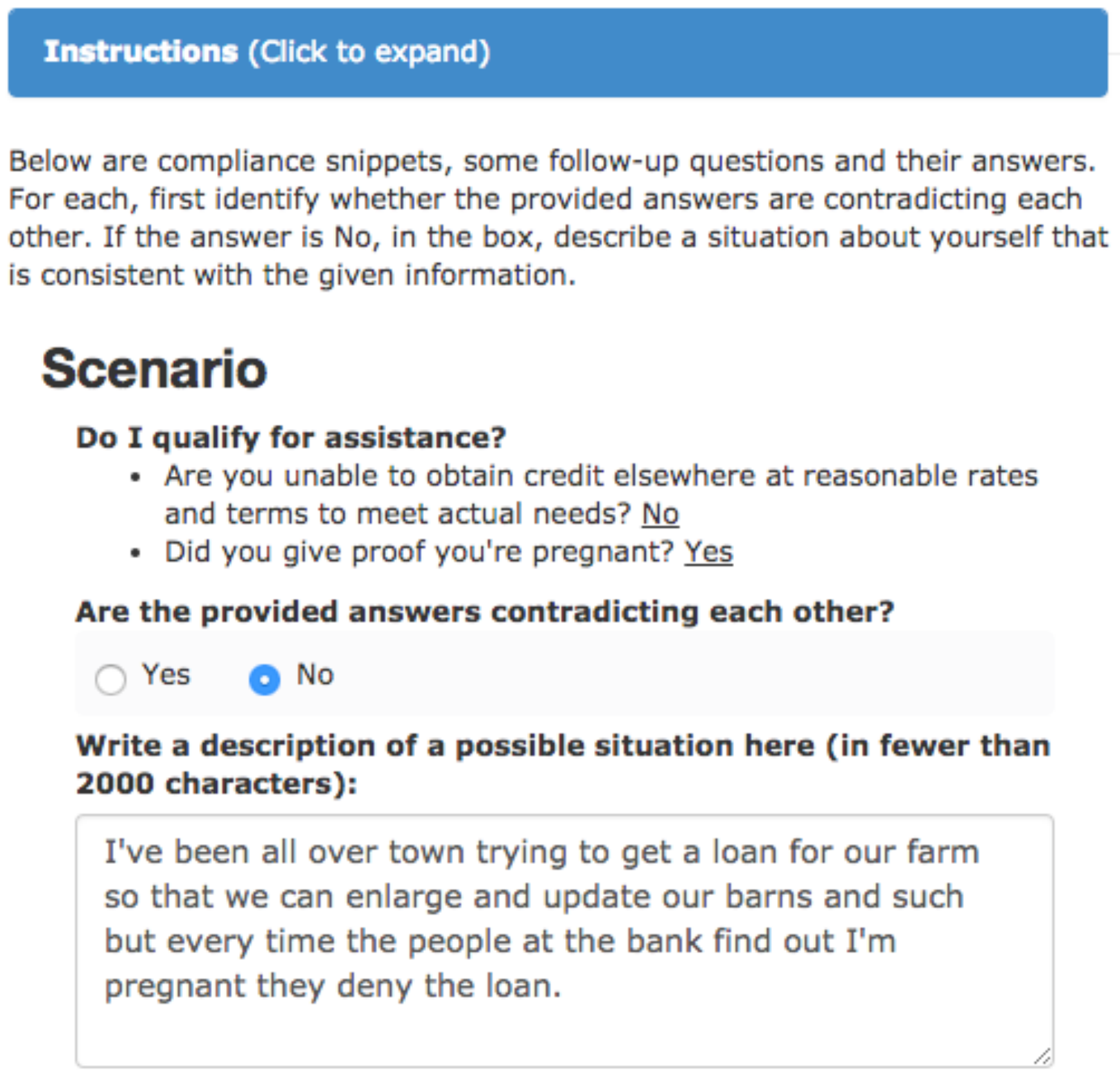}
\caption{\label{scenario_interface} Annotators are asked to write a scenario that fits the given information, i.e. questions and answers.}
\end{figure}

Figure~\ref{scenario_interface} shows the annotation interface for the scenario generation task, where the first question is relevant and the second question is not relevant.

\section{Quality Control}\label{appdx_quality_control}
In this section, we present several measure that we take in order to create a high quality dataset.

\paragraph{Irregularity Detection} A convenient property of the formulation of the reasoning process \SebMargin{I think we need to remove mentions of reasoning process as we haven't introduced it} as a binary decision tree is class exclusivity at the final partitioning of the utterance space\SebMargin{I can't parse this sentence}. That is, if the two leaf nodes stemming from the same \followupquestion{} node have identical \yes{} or \no{} values, this is an indication of either a mis-annotation or a redundant question. 
We automatically identify these irregularities, trim the subtree at \followupquestion{} node and re-annotate. \SebMargin{Refer back to the tree section here?}
This also means that our protocol effectively guarantees a minimum of two annotations per leaf node, further enhancing data quality.

\paragraph{Back-validation} We implement back-validation by providing the workers with two options: \yes{} and proceed with the task, or \no{} and provide an invalidation reason to de-incentivize unnecessary rejections. We found this approach to be valuable both as a validation mechanism as well as a means of collecting direct feedback about the task and the types of incorrect annotations encountered. We then trim any invalidated subtrees and re-annotate.\SebMargin{Again refer to the tree section to make this easier to understand}

\paragraph{Contradiction Detection}
We can introduce contradictory information by adding random questions and answers to a dialog part when generating HITs for scenario generation. 
% For example when the dialog includes the question ``Are you 16 or over? \no{}'' and we add the random question ``Are you over 25 and don't have children? \yes{}'' a scenario cannot be generated that fits all the information. 
Therefore, we first ask each annotator to identify whether the provided dialog parts are contradictory. If they are, the annotator will invalidate the HIT.

\paragraph{Validation Sampling} 
% We use the following metric to detect good workers quickly while minimizing the cost of unnecessary annotation. 
% For each worker with a total annotation count $c_t$, we sample $c_s$ annotations to validate governed by $c_s = \text{ceil}(log_{10}(c_t) + 1)$ for convenience since it satisfies the above requirements.
We sample a proportion of each worker`s annotations to validate. Through this process, each worker is assigned a quality score. We only allow workers with a score higher than a certain value to participate in our HITs \cite{snow2008cheap}. We also restrict participation to workers with $>97\%$ approval rate, $> 1000$ previously completed HITs and located in the UK, US or Canada. 

\paragraph{Qualification Test} Amazon Mechanical Turk allows the creation of qualification tests through the API, which need to be passed by each turker before attempting any HIT from a specific task. A qualification can contain several questions with each having a value. The qualification requirement for a HIT can specify that the total value must be over a specific threshold for the turker to obtain that qualification. We set this threshold to 100\%. 

\paragraph{Possible Sources of Noise}
\Patrick{Max, Marzieh, please review this paragraph}
Here we detail possible sources of noise, estimate their effects and outline the steps taken to mitigate these sources:

a) Noise arising from annotation errors: This has been discussed in detail above.

b) Noise arising from negative question generation: Some noise could be introduced due to the automatic sampling of the negative questions. To obtain an estimate, 100 negative questions were assessed by an expert annotator. It was found that only 8\% of negatively sampled questions were erroneous.

c) Noise arising from the negative scenario sampling: A further 100 utterances with negatively sampled scenarios were curated by an expert annotator, and it was found that 5\% of the utterances were erroneous.

d) Errors arising from the application of scenarios to dialog trees: The assumption that the scenario was only relevant to the follow-up questions it was generated from, and was independent to all other follow-up questions posed in that dialog tree is not necessarily true, and could result in noisy dialog utterances. 100 utterances from the subset of the data where this type of error was possible were assessed by expert annotators, and 12\% of these utterances were found to be erroneous. This type of error can only affect 80\% of utterances, thus the estimated total effect of this type of noise is 10\%. 

Despite the relatively low levels of noise, we asked expert annotators to manually inspect and curate (if necessary) all the instances in the development and the test set that are prone to potential errors. This leads to an even higher quality of data in our dataset.

\MikeMargin{Orphaned lines}

% \clearpage
\section{Further Details on Corpus}\label{appdx_source}
We use $264$ unique sources from $10$ unique domains listed below. For transparency and reproducibility, the source URLs are included in the corpus for each dialog utterance.

\begin{itemize}
  \item \url{http://legislature.maine.gov/}
  \item \url{https://esd.wa.gov/}
  \item \url{https://www.benefits.gov/}
  \item \url{https://www.dmv.org/}
  \item \url{https://www.doh.wa.gov/}
  \item \url{https://www.gov.uk/}
  \item \url{https://www.humanservices.gov.au/}
  \item \url{https://www.irs.gov/}
  \item \url{https://www.usa.gov/}
  \item \url{https://www.uscis.gov/}
\end{itemize}

Further, the \dsacr{} dataset composition can be seen in Table~\ref{table_dataset_breakdown}.
\begin{table}[h]
\centering
\resizebox{\columnwidth}{!}{%
\begin{tabular} { l r r r r }
\toprule
Set & \# Utterances & \# Trees & \# Scenarios & \# Sources \\
\midrule
All & 32436 & 948 & 6637 & 264 \\
Train & 21890 & 628 & 4611 & 181 \\
Development & 2270 & 69 & 547 & 24 \\
Test & 8276 & 251 & 1910 & 59 \\
\bottomrule
\end{tabular}
}
\caption{\label{table_dataset_breakdown}Dataset composition.}

\end{table}

% \clearpage
\section{Negative Data}\label{appdx_negative_data}
In this section, we provide further details regarding the generation of the negative examples.
\subsection{Negative Questions}
\MaxMargin{We should consider addressing Reviewer2 WA1 in anticipation in this section}
Formally, for each unique positive question, rule text pair, $\left(\question_i, \support_i \right)$, and defining $d_i$ as the source document for $\left(\question_i, \support_i \right)$, we construct the set $Q \in \{ \question_1 \ldots \question_n \}$ where $Q$ is the set of questions that are not sourced from $d_i$. We take a random uniform sample $\question_j$ from $Q$ to generate the negative utterance $\left(\question_j, \support_i, \history_j, \answer_j \right)$  where $\answer_j$ = \irrelevant{} and $\history_j$ is an empty history sequence. An example of a negative question is shown below.

\paragraph{Q.} Can I get Working Tax Credit?
\paragraph{R.} You must also wear protective headgear if you are using a learner's permit or are within 1 year of obtaining a motorcycle license.

\subsection{Negative Scenarios}
We also negatively sample scenarios so that models can learn to ignore distracting scenario information that is not relevant to the task. We define a negative scenario as a scenario that provides no information to assist answering a given question and as such, good models should ignore all details within these scenarios. 

A scenario $\scenario_{x}$ is associated with the (one or more) dialog question and answer pairs $\{\left(f_{x,1}, a_{x,1}\right) .. \left(f_{x,n},a_{x,n}\right) \}$ that it was generated from.

For a given unique question, rule text pair, $\left(\question_i, \support_i \right)$, associated with a set of positive scenarios $\{\scenario_{i,1} \ldots \scenario_{i,k}\}$, we uniformly randomly sample a candidate negative scenario $\scenario_j$ from the set of all possible scenarios. We then build TF-IDF representations for the set of all dialog questions associated with $\left(\question_i, \support_i \right)$, i.e. $F_i = \{\left(f_{i,1,1}\right) .. \left(f_{i,k,n}\right) \}$. We also construct TF-IDF representations for the set of dialog questions associated with $\scenario_j$, $F_{s_j} = \{\left(f_{j,1}\right) .. \left(f_{j,x}\right) \}$. 

If the cosine similarity for all pairs of dialog questions between $F_i$ and $F_{s_j}$ are less than a \MikeMargin {Should this show the actual threshold? (0.5)} certain threshold, the candidate is accepted as a negative, otherwise a new candidate is sampled and the process is repeated. Then we iterate over all utterances that contain $\left(\question_i, \support_i \right)$ and use the negative scenario to create one more utterance whenever the original utterance has an empty scenario. The threshold value was validated using manual verification. An example is shown below:

\paragraph{R.} You are allowed to make emergency calls to 911, and bluetooth devices can still be used while driving.
\paragraph{S.} The person I'm referring to can no longer take care of their own affairs.

% \clearpage
\section{Challenges}\label{appdx_challenges}
In this section we present a few interesting examples we encountered in order to provide a better understanding of the requirements and challenges of the proposed task.

\subsection{Dialog Generation}
\begin{figure}[h]
\includegraphics[width=0.5\textwidth, angle=0]{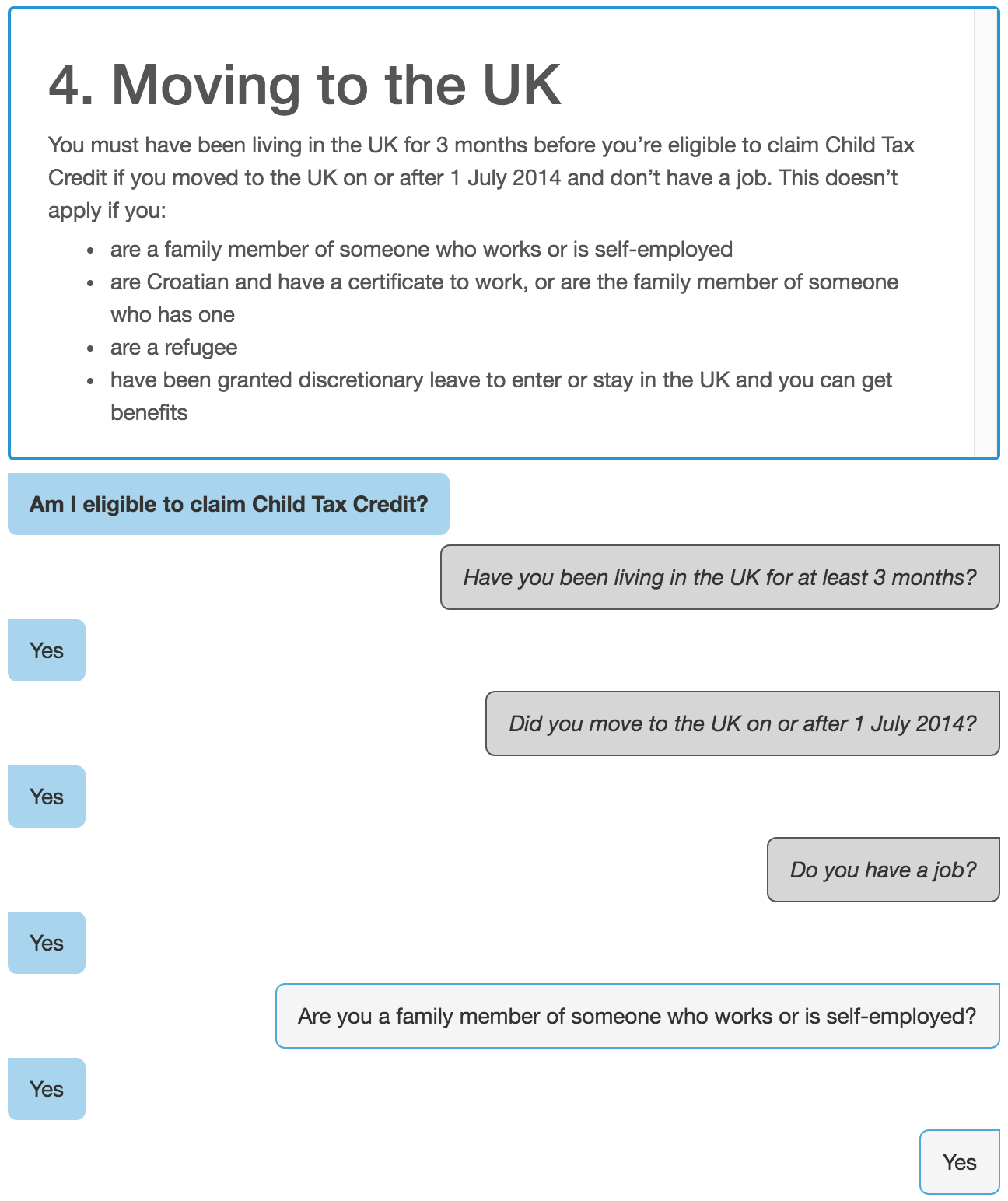}
\caption{\label{complexexample} Example of a complex and hard-to-interpret rule relationship. }
\end{figure}
 
\begin{figure}[h]
\includegraphics[width=0.5\textwidth, angle=0]{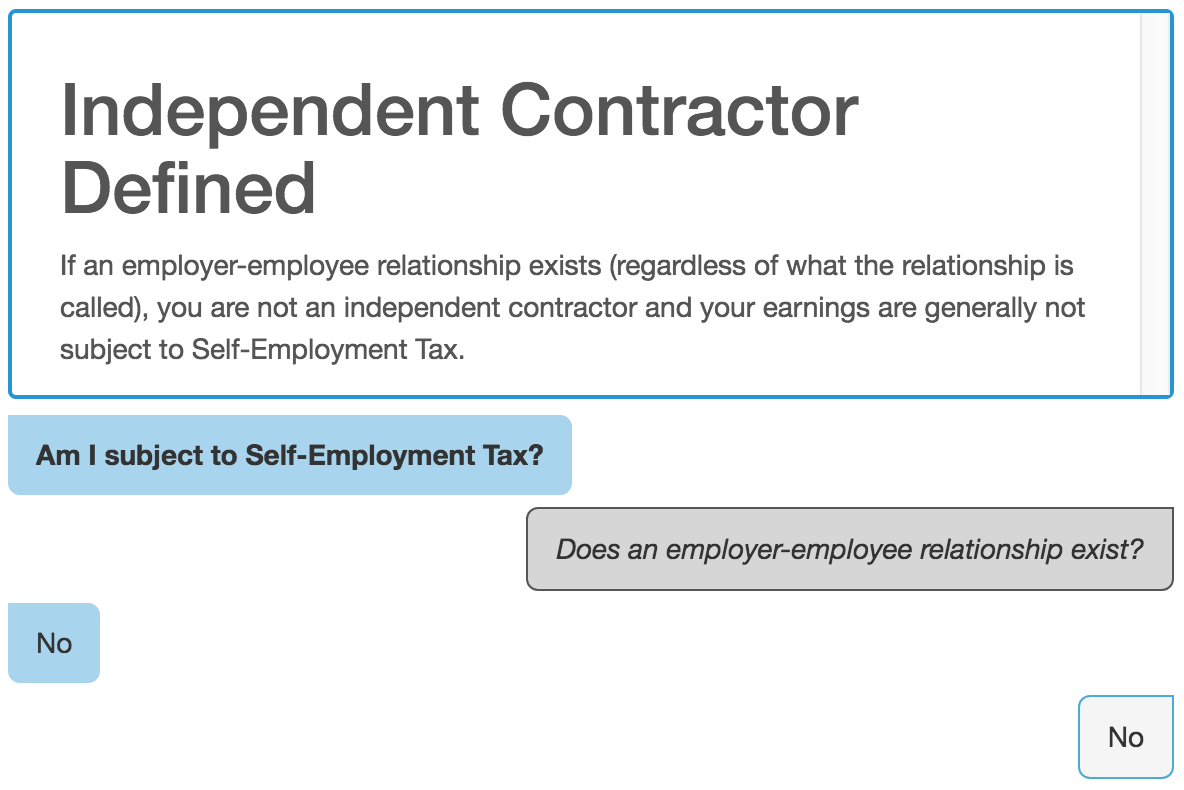}
\caption{\label{example_complexnegation} Example of a hard-to-interpret rule due to complex negations. In this particular example, majority vote was inaccurate. }
\end{figure}

\begin{figure}[h]
\includegraphics[width=0.5\textwidth,angle=0]{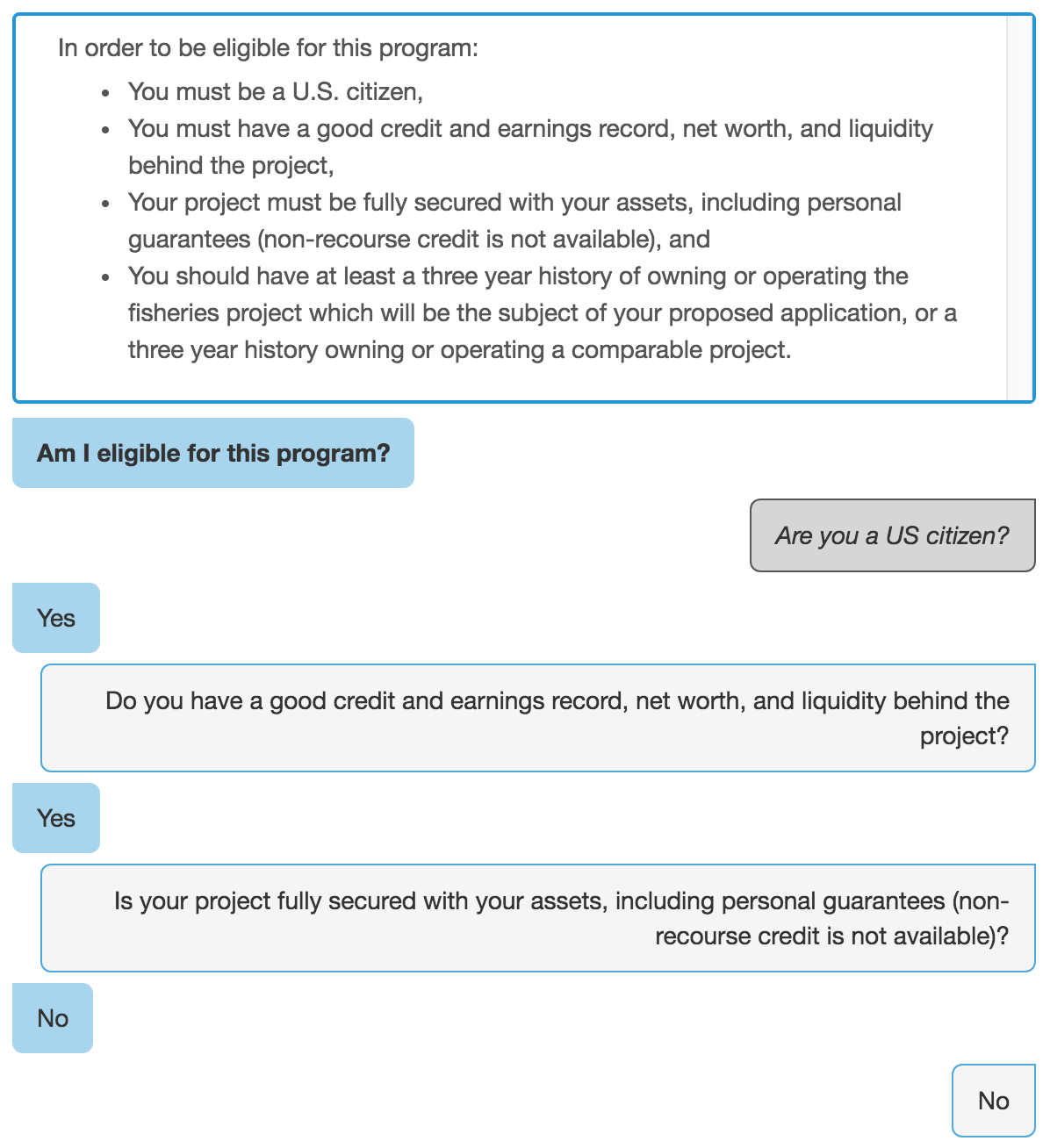}
\caption{\label{example_conjbullets} Example of a conjunctive rule relationship derived from a bulleted list, determined by the presence of ``, and'' in the third bullet. }
\end{figure}

\begin{figure*}[t]
\centering
\includegraphics[width=1\textwidth, angle=0]{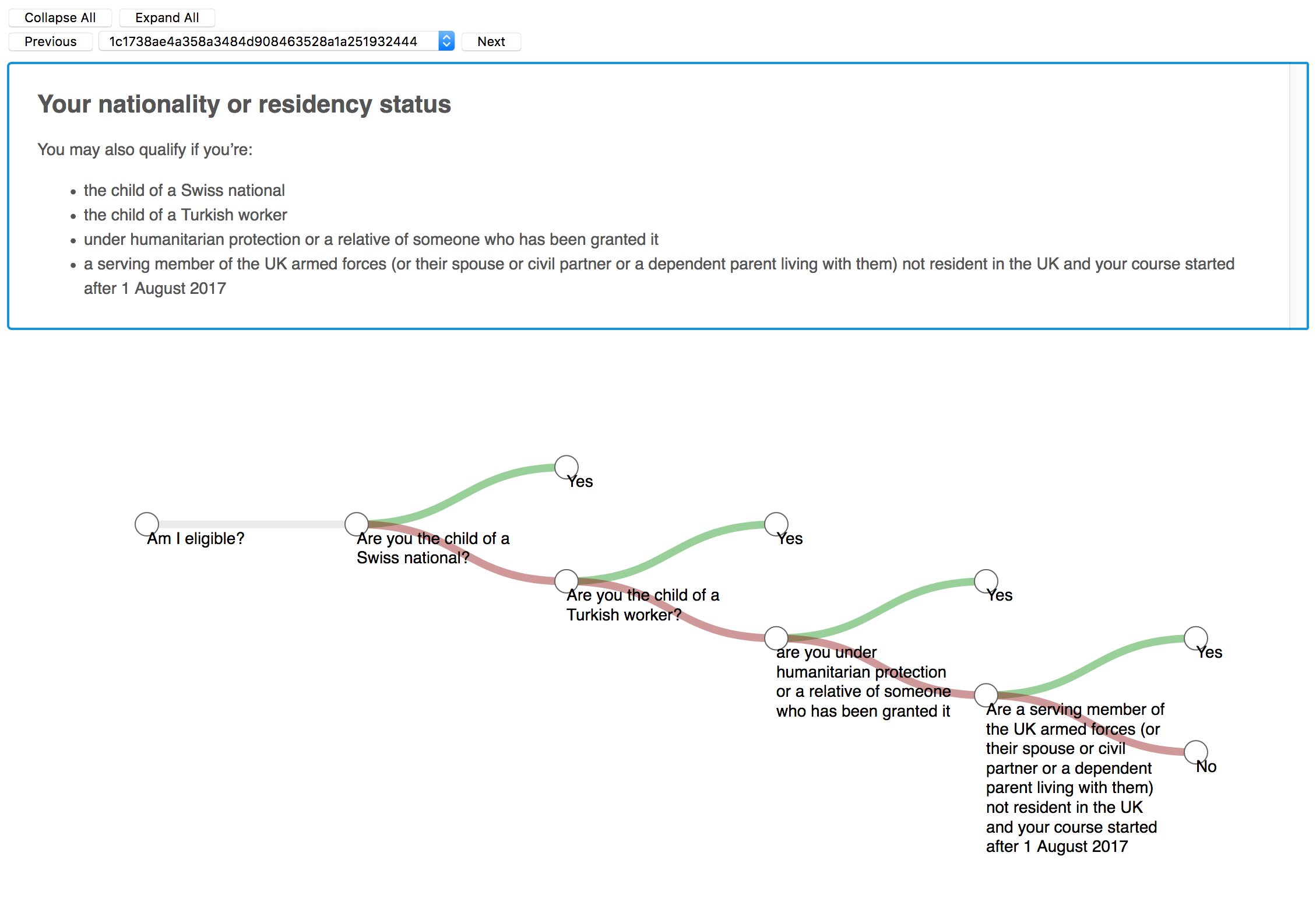}
\caption{\label{example_typicalcascade} Example of a dialog-tree for a typical disjunctive bulleted list. }
\end{figure*}

Table~\ref{table_rule_categories} shows the breakdown of the types of challenges that exist in our dataset for dialog generation and their proportion.

%\begin{figure}[h]
%\includegraphics[width=0.5\textwidth]{img/followup_answer_reasoning_categories.pdf}
%\caption{\label{challenges_followup_answers_reasoning} Percentage of different types of reasoning needed to ask follow-up questions and drive a final answer.}
%\end{figure}

% \clearpage
% \section{Dataset Phenomena}
% \begin{figure*}[t!]
% \centering
% \includegraphics[width=0.8\textwidth, angle=0]{data_analysis.pdf}
% \caption{\label{fig_data_analysis} Histograms of the number and the length of follow-ups (tokens). On the right, the relation between the length of supporting documents (tokens) and the the number of follow-ups bucketed into 25 tokens intervals. The size of circles indicate the number of supporting documents in each bin. }
% \end{figure*}

% \subsection{Scenario Generation}
% 
% \begin{figure}[h]
% \includegraphics[width=0.5\textwidth]{img/reasoning_categories.pdf}
% \caption{\label{challenges_scenario_reasoning} Percentage of different types of reasoning needed to resolve entailment on scenarios and follow-up questions. To see examples, refer to the Table ~\ref{table_scenario_categories}.}
% \end{figure}

% \begin{figure}
% \includegraphics[width=0.45\textwidth]
% {img/scenarios_per_tree.pdf}
% \caption{\label{scenarios_histogram} Histogram showing the number of scenarios that have been generated for each tree.}
% \end{figure}

% \clearpage
\section{Entailment Corpus}\label{appdx_entailment_corpus}
Using the scenarios and their associated questions and answers we create an entailment corpus for each of the train, development and test sets of \dsacr{}. For every dialog utterance that includes a scenario, we create a number of data points as follows:

For every utterance in \dsacr{} with input $x = (\question, \support, \history, \scenario)$ and output $\answer$ where $\answer = f_m \not \in \{\yes, \no, \irrelevant \} $, we create an entailment instance ($x_e$, $y_e$) such that $x_e$ = \scenario \\ and:

\begin{itemize}
\item $y_e$ = \entailment{} if the answer $a_m$ to follow-up question $f_m$ is \yes{} which can be derived from \scenario.
\item $y_e$ = \contradiction{} if the answer $a_m$ to follow-up question $f_m$ is \no{} which can be derived from \scenario.
\item $y_e$ = \neutral{} if the answer $a_m$ to follow-up question $f_m$ cannot be derived from \scenario.
\end{itemize}
Table~\ref{emtailment_stat} shows the statistics for the entailment corpus.

\begin{table}[h]
\centering
\small
\begin{tabular}{l | r  r r}
\toprule
Set & \entailment{} & \contradiction{} & \neutral{} \\
\midrule
Train & $2373$ & $2296$ & $10912$\\
Dev & $271$ & $253$ & $1098$\\
Test & $919$ & $944$ & $4003$ \\
\bottomrule
\end{tabular}
\caption{\label{emtailment_stat}Statistics of the entailment corpus created from the \dsacr{} dataset.}
\end{table}

\clearpage
\onecolumn
\section{Further details on Interpreting rules}\label{appdx_interpreting_rules}

\begin{table*}[htb]
\small
\resizebox{\linewidth}{!}{
\begin{tabular}{p{25mm} p{30mm} p{75mm} r }
\toprule
\textbf{Category} & \textbf{Example Question} & \textbf{Example Rule Text} & \textbf{Percentage} \\
\hline
\midrule
Simple & Can I claim extra MBS items? & If you’re providing a bulk billed service to a patient you may claim extra MBS items. & 31\% \\
\midrule
Bullet Points & Do I qualify for assistance? & To qualify for assistance, applicants must meet all loan eligibility requirements including:
\begin{itemize}[itemsep=0pt,parsep=2pt,topsep=5pt]
\item Be unable to obtain credit elsewhere at reasonable rates and terms to meet actual needs;
\item Possess legal capacity to incur loan obligations;
\end{itemize}
 & 34\% \\
\midrule
In-line Conditions & Do these benefits apply to me? &These are benefits that apply to individuals who have earned enough Social Security credits and are at least age 62. & 39\% \\
\midrule
Conjunctions & Could I qualify for Letting Relief? & If you qualify for Private Residence Relief and have a chargeable gain, you may also qualify for Letting Relief. This means you’ll pay less or no tax. & 18\% \\
\midrule
Disjunctions & Can I get deported? & The United States may deport foreign nationals who participate in criminal acts, are a threat to public safety, or violate their visa. & 41\% \\
\midrule
Understanding Questioner Role & Am I eligible? & The borrower must qualify for the portion of the loan used to purchase or refinance a home. Borrowers are not required to qualify on the portion of the loan used for making energy-efficient upgrades. & 10\% \\
\midrule
Negations & Will I get the National Minimum Wage? & You won’t get the National Minimum Wage or National Living Wage if you’re work shadowing & 15\% \\
\midrule
Conjunction Disjunction Combination & Can my partner and I claim working tax credit? & You can claim if you work less than 24 hours a week between you and one of the following applies:
\begin{itemize}[itemsep=0pt,parsep=2pt,topsep=5pt]
\item you work at least 16 hours a week and you’re disabled or aged 60 or above 
\item you work at least 16 hours a week and your partner is incapacitated
\end{itemize} & 18\% \\
\midrule
World Knowledge Required to Resolve Ambiguity & Do I qualify for Statutory Maternity Leave? & You qualify for Statutory Maternity Leave if:
\begin{itemize}[itemsep=0pt,parsep=2pt,topsep=5pt]
\item you’re an employee not a `worker'
\item you give your employer the correct notice
\end{itemize} & 13\% \\
\bottomrule
\end{tabular}
}
\caption{Types of features present for question, rule text pairs and their proportions in the dataset based on 100 samples. World Knowledge Required to resolve ambiguity refers to where the rule itself doesn't syntactically indicate whether to apply a conjunction or disjunction, and world knowledge is required to infer the rule.}
\label{table_rule_categories}
\end{table*}

% \clearpage
\twocolumn
\section{Further details on Follow-up Question Generation Modelling}\label{appdx_question_gen}
Table \ref{followup_results_table_all} details all the results for all the the models considered for follow-up question generation.

\paragraph{First Sent.} Return the first sentence of the rule text
\paragraph{Random Sent.} Return a random sentence from the rule text
\paragraph{SurfaceLR} A simple binary logistic model, which was trained to predict whether or not a given sentence in a rule text had the highest trigram overlap with the target follow-up question, using a bag of words feature set, augmented with 3 very simple engineered features (the number of sentences in the rule text, the number of tokens in the sentence and the position of the sentence in the rule text)

\paragraph{Sequence Tag} A simple neural model consisting of a learnt word embedding followed by an LSTM. Each word in the rule text is classified as either in or out of the subsequence to return using an I/O sequence tagging scheme. 

\begin{table*}[t]
\small
\centering
\begin{tabular}{l | c | c | c | c }
\toprule
Model & BLEU-1 & BLEU-2 & BLEU-3 & BLEU-4 \\
\midrule
Random Sent.& $0.302$ & $0.228$ & $0.197$ & $0.179$\\
First Sent.& $0.221$ & $0.144$ & $0.119	$ & $0.106$ \\
Last Sent.& $0.314$ & $0.247$ & $0.217$ & $0.197$ \\
Surface LR & $0.293$ & $0.233$ & $0.205$ & $0.186$ \\
NMT-Copy & $0.339$ & $0.206$ & $0.139$ & $0.102$\\
Sequence Tag & $0.212$&$0.151$&$0.126$&$0.110$\\
BiDAF & $0.450$&$0.375$&$0.338$&$0.312$\\
Rule-based & $\mathbf{0.533}$&$\mathbf{0.437}$&$\mathbf{0.379}$&$\mathbf{0.344}$\\
\bottomrule
\end{tabular}
\caption{All results of the baseline models on follow-up question generation.}
\label{followup_results_table_all}
\end{table*}

\subsection{Further details on neural models for question generation}

Table \ref{neural_encoding_table_for_followups} details what the inputs and outputs of the neural models should be.
\begin{table*}[t]
\scriptsize
\centering
\begin{tabular}{l | p{5cm} | p{5cm}  }
\toprule
Model & Input & Output \\
\midrule
NMT-Copy & \support{}  $||$ \question{}  $||$ $f_1$ ? $a_a$ $||$ \ldots $||$ $f_m$ ? $a_m$ & $f_{m+1}$\\
Sequence Tag & \support{}  $||$ \question{}  $||$ $f_1$ ? $a_a$ $||$ \ldots $||$ $f_m$ ? $a_m$ & Span corresponding to follow-up question.\\
BiDAF & Question: \question{} $||$ $f_1$ ? $a_a$ $||$ \ldots $||$ $f_m$ ? $a_m$ \newline Context \ : \ \support{} & Span corresponding to follow-up question. \\
\bottomrule
\end{tabular}
\caption{Inputs and outputs of neural models for question generation.}
\label{neural_encoding_table_for_followups}
\end{table*}

The NMT-Copy model follows an encoder-decoder architecture. The encoder is an LSTM. The decoder is a GRU equipped with a copy mechanism, with an attention mechanism over the encoder outputs and an additional attention over the encoder outputs with respect to the previously copied token. We achieved best results by limiting the model`s generator vocabulary to only very common interrogative words. We train with a 50:50 teacher-forcing / greedy decoding ratio. At test time we greedily sample the next word to generate, but prevent repeated tokens being generated by sampling the second highest scoring token if the highest would result in a repeat.

In order to frame the task as a span extraction task, a simple method of mapping a follow-up question onto a span in the rule text was employed. The longest common subsequence of tokens between the rule text and follow-up question was found, and if the subsequence length was greater than a certain threshold, the target span was generated by increasing the length of the subsequence so that it matched the length of the follow-up question. These spans were then used to supervise the training of the BiDAF and sequence tagger models. 

% \clearpage
\section{Evaluating Utility of CMR}
\label{app:e2e}

In order to evaluate the utility of conversational machine reading, we run a user study that compares CMR with the scenario when such an agent is not available, i.e. the user has to read the rule text, the question, and the scenario, and determine for themselves whether the answer to the question is \quoteYes{} or \quoteNo{}.
On the other hand, with the agent, the user does not read the rule text, instead only responds to follow-up questions with a \quoteYes{} or \quoteNo{}, based on the scenario text and world knowledge.

% \paragraph{Setup}
We carry out a user study with 100 randomly selected scenarios and questions, and elicit annotation from $5$ workers for each.
As these instances are from the CMR dataset, the quality is fairly high, and thus we have access to the \emph{gold} answers and follow-ups questions for all possible responses by the users.
This allows us to evaluate the accuracy of the users in answering the question, the primary objective of any QA system.
We also track a number of other metrics, such as the time taken by the users to reach the conclusion. % and the number of words read.

\begin{figure}[t]
	\centering
	\begin{subfigure}[b]{\columnwidth}
		\centering
		\includegraphics[width=0.9\textwidth]{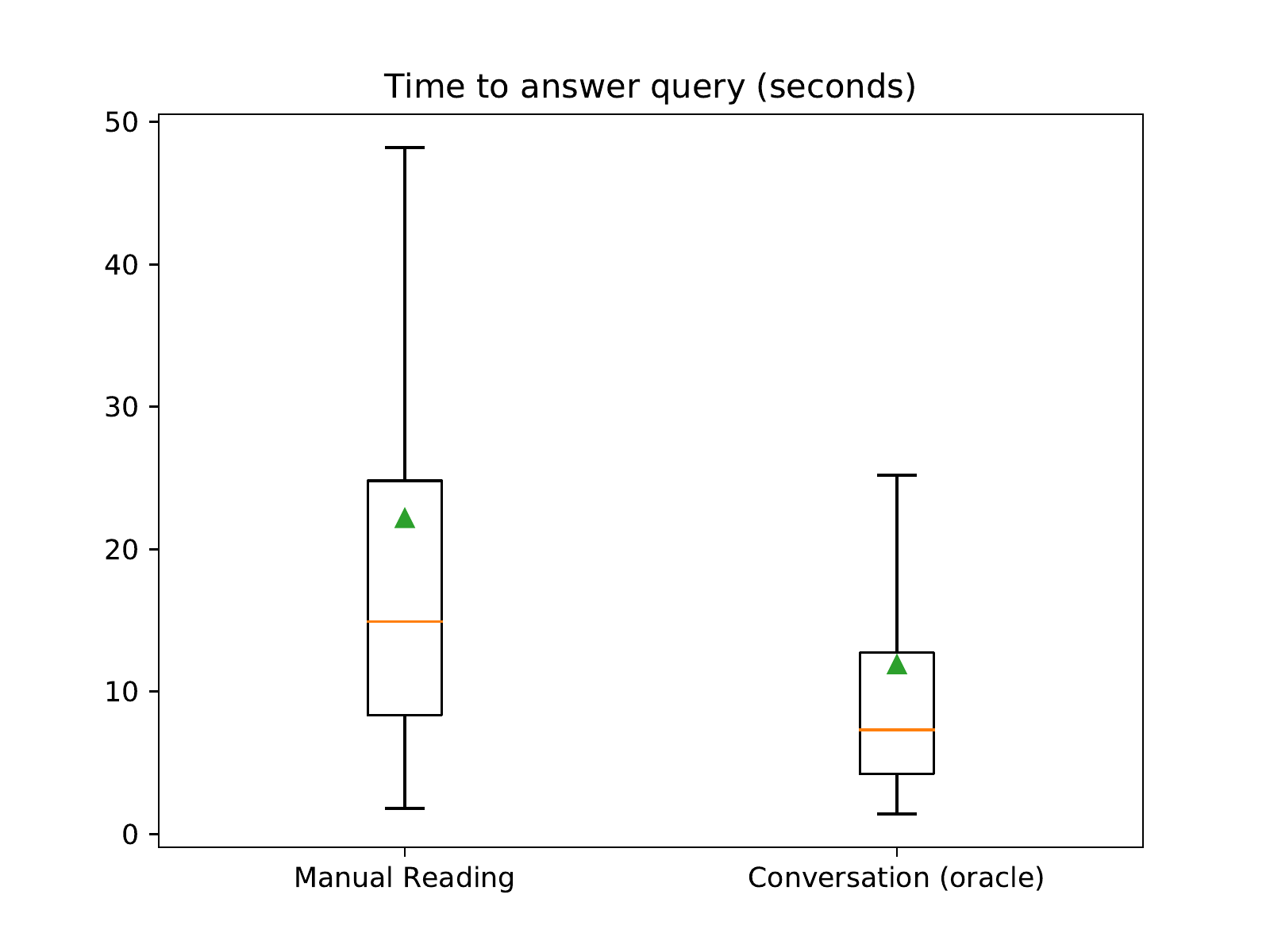}
		
        \caption{\label{fig:e2e:time}Time taken to reach conclusion}
	\end{subfigure}
	\begin{subfigure}[b]{\columnwidth}
		\centering
		\includegraphics[width=0.9\textwidth]{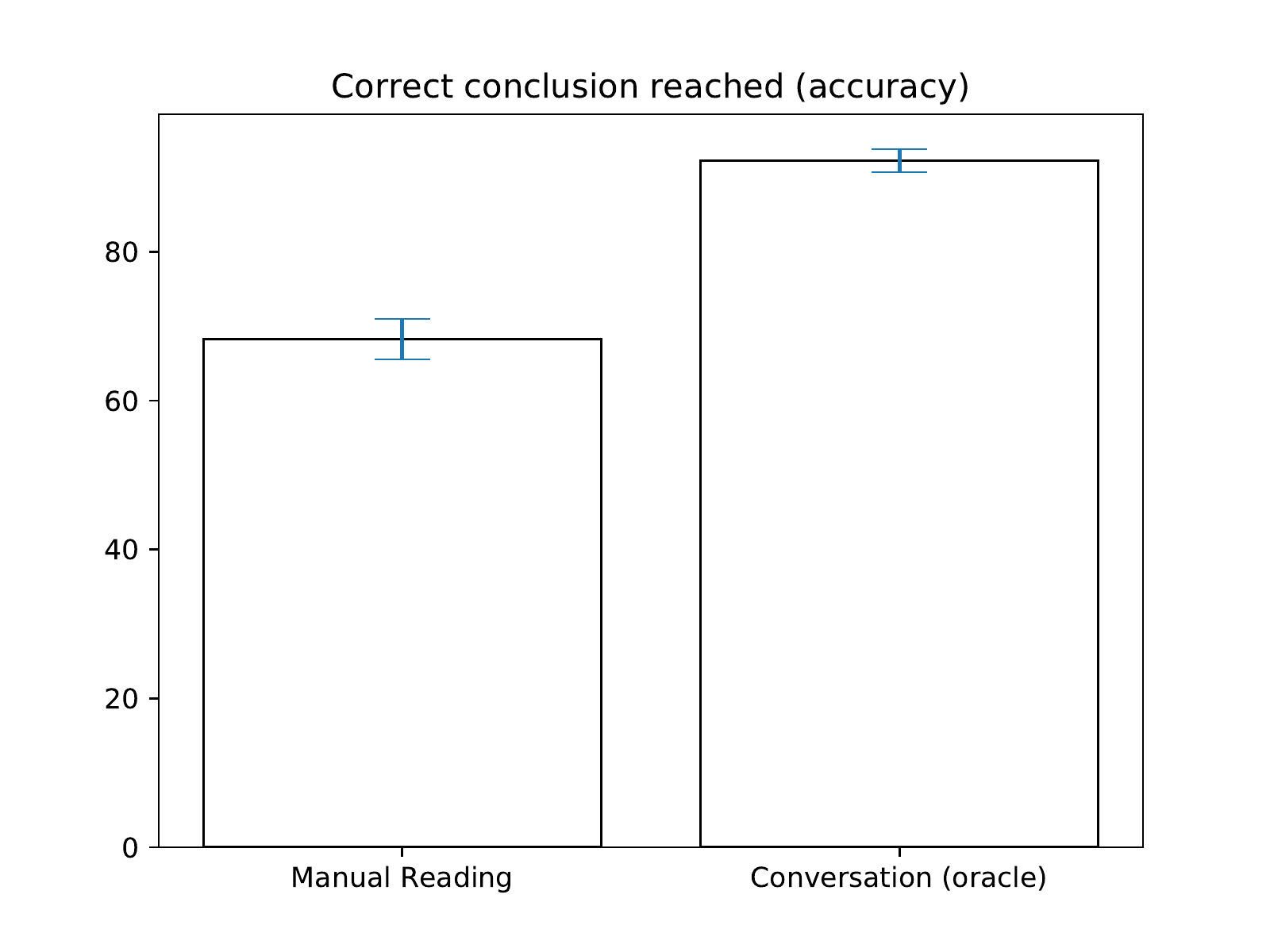}
		
        \caption{\label{fig:e2e:acc}Accuracy of the conclusion reached}
	\end{subfigure}
    
    \caption{\label{fig:e2e}\textbf{Utility of CMR} Evaluation via a user study demonstrating that users with an accurate conversational agent are not only reach conclusions much faster than ones that have to read the rule text, but also that the conclusions reached are correct much more often.}
\end{figure}

% \paragraph{Results}
% The results are shown in Figure~\ref{fig:e2e}.
In Figure~\ref{fig:e2e:time}, we see that the users that have access to the conversational agent are almost twice as fast the users that need to read the rule text.
This demonstrates that even though the users with the conversational agent have to answer more questions (as many as the followup questions), they are able to understand and apply the knowledge more quickly.
Further, in Figure~\ref{fig:e2e:acc}, we see that users with access to the conversational agents are \emph{much more} accurate than ones without, demonstrating that an accurate conversational agent can have a considerable impact on efficiency. % in this setting.

% \vfill
% \vfill

\end{document}